\documentclass{article}

\usepackage[final]{corl_2020} % Uncomment for the camera-ready ``final'' version.

\usepackage{graphicx}
\graphicspath{{figures/}}
\usepackage[export]{adjustbox}
\usepackage{caption}
\usepackage{subcaption}
\usepackage{wrapfig}
\usepackage{algorithm}
\usepackage[noend]{algpseudocode}
\usepackage{setspace}
\usepackage{hyperref}
\usepackage{tikz}
\usetikzlibrary{positioning,shapes}
\usepackage{bm}
\usepackage{xcolor}
\usepackage{siunitx}
\usepackage{balance}
\usepackage{booktabs}
\usepackage{amsmath, amssymb}
\usepackage{subcaption}
\usepackage[percent]{overpic}
\usepackage[textsize=scriptsize,textwidth=1cm]{todonotes}

\title{From Pixels to Legs: \\Hierarchical Learning of Quadruped Locomotion}

\author{
  Deepali Jain \\
  Robotics at Google \\
  \texttt{jaindeepali@google.com}
  \And
  Atil Iscen \\
  Robotics at Google \\
  \texttt{atil@google.com}
  \And
  Ken Caluwaerts \\
  Robotics at Google \\
  \texttt{kencaluwaerts@google.com}
}

\usepackage{subcaption}

\usepackage[textsize=scriptsize,textwidth=1cm]{todonotes}

\newcommand{\deepalisays}[1]{}
\newcommand{\atilsays}[1]{}
\newcommand{\kensays}[1]{}

\begin{document}
\maketitle

\setlength{\abovedisplayskip}{1pt}
\setlength{\belowdisplayskip}{3pt}

%===============================================================================

\begin{abstract}
Legged robots navigating crowded scenes and complex terrains in the real world are required to execute dynamic leg movements while processing visual input for obstacle avoidance and path planning.
We show that a quadruped robot can acquire both of these skills by means of hierarchical reinforcement learning (HRL). \deepalisays{removed end-to-end before HRL. better way to say jointly trained?}
By virtue of their hierarchical structure, our policies learn to implicitly break down this joint problem by concurrently learning High Level (HL) and Low Level (LL) neural network policies. These two levels are connected by a low dimensional hidden layer, which we call latent command. 
HL receives a first-person camera view, whereas LL receives the latent command from HL and the robot's on-board sensors to control its actuators. 
We train policies to walk in two different environments: a curved cliff and a maze. 
We show that hierarchical policies can concurrently learn to locomote and navigate in these environments, and show they are more efficient than non-hierarchical neural network policies. 
This architecture also allows for knowledge reuse across tasks. LL networks trained on one task can be transferred to a new task in a new environment.
Finally HL, which processes camera images, can be evaluated at much lower and varying frequencies compared to LL, thus reducing computation times and bandwidth requirements. Video illustrations of our learned policies are available at this \href{https://drive.google.com/file/d/1zgbY3eEYf1scDFu07JRFhL_cL8rfgN14/view}{link}\footnote{https://rb.gy/jacqsb}.

\end{abstract}


%===============================================================================

\section{Introduction}
Legged robots have the potential to traverse many types of terrains while demonstrating a diverse set of agile skills. However, control of legged robots is challenging due to the dynamic nature of the problem. When incorporating visual inputs in the control loop, the task at hand becomes more difficult as it requires perceiving the environment  while simultaneously handling the fast and contact-rich dynamics of the robot's legs.

One solution is to split the problem into independent modules for vision and dynamics. However, this approach is typically limited by the high-level features and low-level behaviors that are independently designed or learned. It may be impossible to come up with a single feature space that is optimal for all given tasks, or the low-level behaviors needed might be different for each given task. We  tackle the two problems of vision processing and fast dynamics by designing a hierarchical architecture with a high level (HL) and low level (LL) subsystem, which are concurrently trained. This framework does not require design decisions beyond a standard RL setup. HL handles vision with variable frequency and outputs a latent command, which is passed on to LL. LL runs at a higher frequency and handles control of the legs. 

We build on the hierarchical architecture presented in a related work by Jain et al~\cite{Jain2019HierarchicalRL} and incorporate vision processing to learn to navigate environments while concurrently discovering legged locomotion skills. The architecture is trained using Evolutionary Strategies (ES)~\cite{salimans2017evolution}. The main contributions of our research are as follows:
\begin{itemize}
    \item Our HRL solution implicitly learns a complete pipeline from pixels to motor commands for quadruped legged locomotion without the need to design or learn low-level behaviors.
    We show that a hierarchical policy with more than $10^5$ parameters can be successfully trained using evolutionary strategies. 
    \item Separation of observations into hierarchical levels allows knowledge reuse, because behaviors learned by LL are largely task agnostic and transferable to tasks of a similar nature. We show that data efficiency can be further improved by transferring LL from previously solved tasks, even if they were trained in a different environment. 
    \item The high level runs at a variable frequency computed by the high level's neural network. The result is that visual inputs are processed at much lower frequency ($\SI{1.5}{\hertz}-\SI{10}{\hertz}$) compared to the low level ($\SI{500}{\hertz}$). This leads to more efficient exploration and reduced training times, as illustrated by our experiments.
    \item We illustrate how locomotion primitives emerge and can be selected as a low dimensional latent command. This creates an information bottleneck in which HL learns to extract only the useful visual information and LL learns only primitives relevant to the environment and the task at hand. We provide a detailed analysis of these specific learned behaviors.
\end{itemize}

To test our method, we use a highly realistic simulation model of the Laikago robot, a quadruped with $12$ degrees of freedom. The model is created in PyBullet~\cite{pybulletcoumans} software and carefully tuned based on the physical robot. Despite our temporary inability to validate the results on hardware, we are confident in transferring our learned policies. In prior work, we have demonstrated successful deployment of policies learned in simulation on the robot. Additionally, we have performed validation experiments for the HL by processing real world depth images and verifying the computed latent commands. More details are provided in Appendix~\ref{sec:robot_transfer}. 
We test our framework on $3$ visual navigation tasks and compare our policies with a non-hierarchical baseline. Our method outperforms the baseline and achieves increased sample efficiency and a lower wall-clock training time. Furthermore, we show that by running HL at low frequency, the inference time and computation cost of the learned policy is reduced with minimal effect on task performance.

\section{Background and Related Work}
\textbf{Hierarchical Reinforcement Learning.} HRL decomposes complex decision making into sub-problems. Well-known HRL frameworks in literature are based on Options~\cite{sutton1999between}, MAXQ value decomposition~\cite{dietterich2000hierarchical} and Hierarchical Abstract Machines~\cite{parr1998reinforcement}. In these frameworks, a HL policy typically outputs temporally extended actions which are executed by LL for a specified amount of time. 
Designing or training good LL policies quickly becomes challenging for complicated tasks such as those encountered in robotics.
Some papers try to learn LL by imitating from reference data~\cite{merel2018hierarchical, Peng2019MCPLC, peng2017deeploco}. This approach relies heavily on high quality data-sets, which are often hard to obtain. 

Recognizing the challenges of pre-training LL, we focus on a framework for learning both levels concurrently from scratch; many approaches have been proposed for this. 
In one class of methods, a sub-goal conditioned LL is trained to reach a point in observation space specified by HL~\cite{Li2019LearningGL, levy2018hierarchical, nachum2018data}. However, this interface is not suitable in the case of high-dimensional observation spaces, such as camera images~\cite{nachum2018near}.
Some methods design auxiliary rewards to promote diversity in LL skills~\cite{hausman2018learning, florensa2017stochastic, CoReyes2018SelfConsistentTA}; however, by doing so, the RL agent may be forced to learn many skills irrelevant to the given task, leading to inefficiency in learning. Bacon et al~\cite{bacon2017option} use an intrinsic reward function to learn LL. On the contrary, by training our policy with gradient-free policy search, we are able to train the whole hierarchical architecture from the main task reward. Thus we avoid imposing any external priors on LL behavior through intrinsic or auxiliary rewards. 
Some methods learn a finite set of options~\cite{bacon2017option, fox2017multi, daniel2013learning}. In our solution, we adopt the hierarchical policy structure proposed in \cite{Jain2019HierarchicalRL}, which uses a vector space for modulating LL, allowing it to learn a continuum of skills. This is required to solve agile locomotion tasks.

\textbf{Legged Locomotion.}  Reinforcement Learning (RL) has been successfully applied to the problem of learning basic locomotion skills~\cite{schulman2017proximal, iscen2018policies, hwangbo2019learning}. Complex locomotion tasks have also been addressed using RL~\cite{peng2017deeploco, peng2016terrain, merel2018hierarchical, Heess2016LearningAT, xie2019iterative}. Often, as complexity grows, only a part of the pipeline uses RL. Domain knowledge is used to constrain the problem, usually through a hand-crafted hierarchical solution~\cite{peng2017deeploco, Li2019LearningGL}. This is especially true for locomotion on real robots~\cite{hwangbo2019learning}.
Peng et al~\cite{peng2017deeploco} use HRL to learn locomotion in physics-based character animation. The two levels are learned separately by means of RL.
Li et al~\cite{Li2019LearningGL} learn locomotion on the hexapod robot named Daisy. In their solution, HL does model planning to select one out of a few pre-trained LL skills. In this work, we solve vision-based legged locomotion without designing or pre-training LL.

\textbf{Perception in Robotics.} Solving robotics tasks from vision input is an important and well-researched topic~\cite{kalashnikov2018qt, yahya2017collective, levine2016end, Pan2019ZeroshotIL}. 
Our work focuses on learning legged locomotion and necessary navigation skills from vision. Prior work has considered HL kinematic or wheeled navigation from vision~\cite{Pan2019ZeroshotIL, Li2019HRL4INHR, blanc2005indoor}. Our method learns LL legged locomotion directly from vision input, which involves processing vision to obtain navigation directives and learning dynamic legged locomotion skills to execute those directives.
Some solutions for vision-based legged locomotion have been proposed that use domain-specific, hand-designed pipelines~\cite{bazeille2013vision, magana2019fast}.

\textbf{Evolutionary Strategies in Robotics.} Many RL algorithms for continuous control are available for training policies to solve robotics tasks; policy gradient and actor-critic methods are especially popular. However, given the architecture of our hierarchical policy, training with derivative-free approaches is the reasonable choice. Recently, evolutionary methods~\cite{salimans2017evolution} have been successfully applied in robotics~\cite{iscen2018policies, gao2020robotic, choromanski2020provably, song2019reinforcement}. We use an evolutionary algorithm called Augmented Random Search (ARS)~\cite{mania2018simple} to optimize our neural network policies.

\section{Method}
\label{sec:method}

\begin{figure}[h]
\centering
\begin{minipage}{0.6\linewidth}
\includegraphics[width=1\linewidth]{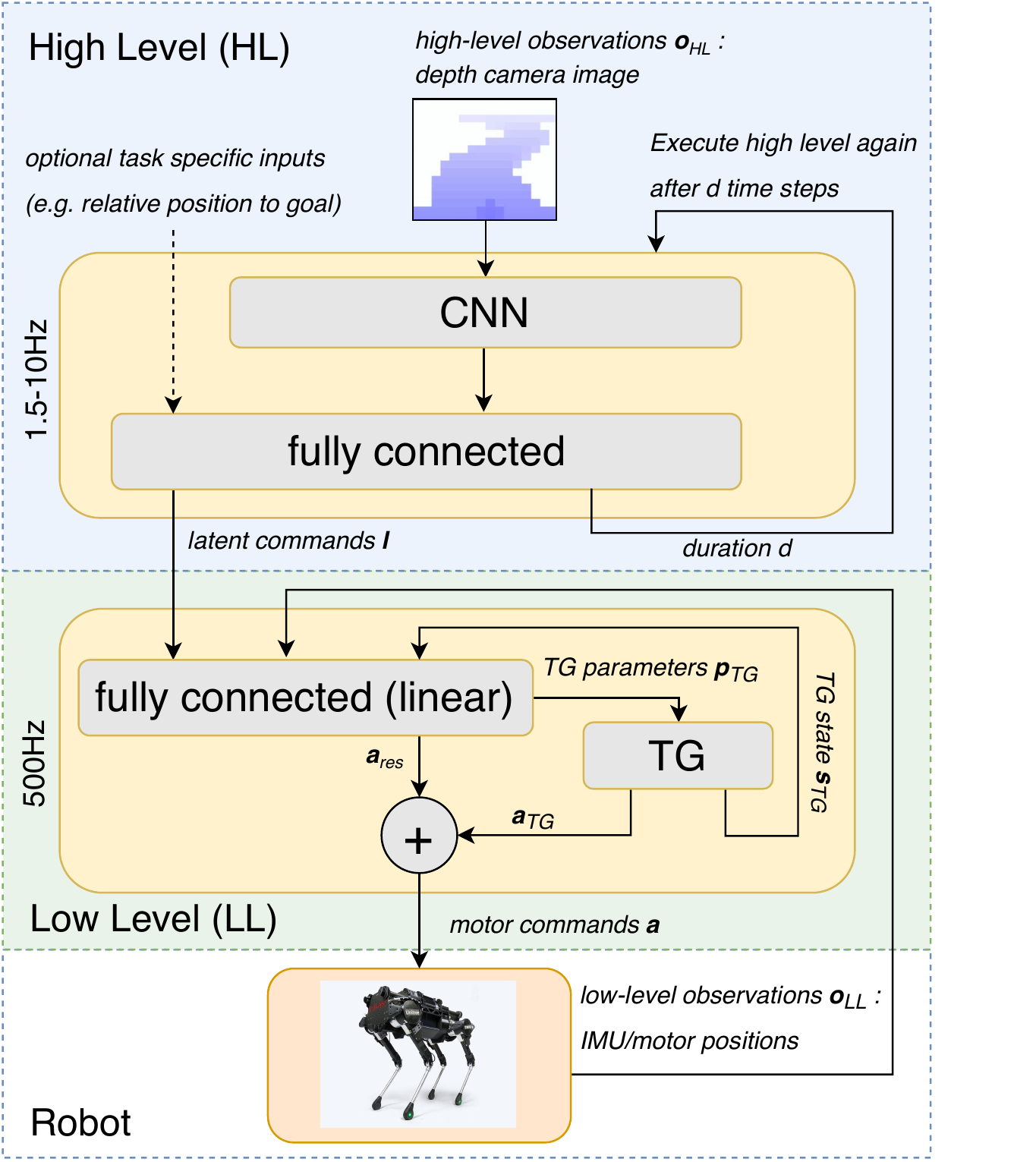}
\end{minipage}%
\begin{minipage}{0.4\linewidth}
\captionof{figure}{\textbf{Hierarchical policy.} The high level (HL) is a CNN with parameters $\bm{\theta}_{HL}$. The HL receives  depth camera observations $\bm{o}_{HL}$ and outputs a \emph{latent command vector} $\bm{l}$ and a duration $d$. Optionally, task specific inputs can be fed into the HL's fully connected output layer. The low level (LL) is a linear network with parameters  $\bm{\theta}_{LL}$. It computes motor actuation commands $\bm{a}_{res}$ and trajectory generator parameters $\bm{p}_{TG}$ based on $\bm{l}$, trajectory generator state $\bm{s}_{TG}$, and low-level observations $\bm{o}_{LL}$ (IMU sensor values and motor angles). $\bm{a}_{res}$ is added to the motor commands from trajectory generator $\bm{a}_{TG}$ and applied to the robot motors. The HL is only evaluated every $d$ steps. In our experiments, the low level runs at the environment simulator's frequency of $\SI{500}{\hertz}$, while the high level policy runs at $\SI{1.5}{\hertz}-\SI{10}{\hertz}$. The HL and LL level networks are trained concurrently by an evolutionary algorithm.}
\label{fig:hrlPolicy1}
\end{minipage}
\end{figure}

% \begin{figure*}
% \centering
% \includegraphics[width=0.8\linewidth]{latentHRLPolicyDuration.pdf}
% \caption{Hierarchical policy evaluation timeline. The high-level policy computes a latent command for the low-level policy and a duration for which to execute the low-level policy. The low-level policy interacts with the hardware at a constant frequency. At the end of the high-level period, the high-level receives updated high-level observations and computes a new latent command and duration.}
% \label{fig:policy_duration}
% \end{figure*}

\begin{algorithm}
\caption{Executing a Hierarchical Policy}\label{alg:rollout}
\begin{algorithmic}[1]
\Procedure{RunHRLPolicy}{$\bm{\theta}_{HL}, \bm{\theta}_{LL}$}\Comment{\parbox[t]{.4\linewidth}{HRL policy weights}}
 %\State \{$\bm{\theta}_{HL}, \bm{\theta}_{LL} \} = \bm{\Theta}$
 %\State $o_h, o_l \gets$ initial HL and LL observations
%  \State $\bm{o}_h \gets$ initial HL observation
 %\State $d \gets 0$ \Comment{\parbox[t]{.5\linewidth}{LL duration}}
 \State $\{R, d, \bm{s}_{TG}\} \gets \{0, 0, \bm{0}\}$ \Comment{\parbox[t]{.4\linewidth}{Return, HL duration, TG state}}
 %\State $R \gets 0$\Comment{\parbox[t]{.4\linewidth}{Episode cumulative reward}}
 %\State $d \gets 0$ \Comment{\parbox[t]{.4\linewidth}{HL duration}}
 \While{not end of episode}
  \If{$d = 0$}
    %\State $d, \bm{l} = \phi_h \times o_h$ \Comment{\parbox[t]{.5\linewidth}{get LL duration and latent command from HL}}
    \State $\bm{o}_{HL} \gets$ Most recent depth image \Comment{\parbox[t]{.4\linewidth}{HL observation}}
    \State \{$d, \bm{l}\} \gets f_{\bm{\theta}_{HL}}(\bm{o}_{HL})$ \Comment{\parbox[t]{.4\linewidth}{HL execution}}
  \EndIf
%   \State $o_l \gets$ \Call{ExtractLowLevelObservation}{$o_h$}
  %\State $\bm{o}_l = \{\bm{o}_l, \bm{l}\}$ %\Comment{\parbox[t]{.5\linewidth}{append latent command to LL observation\vspace{0.2cm}}}
  \State $\bm{o}_{LL} \gets$ IMU \& motor positions \Comment{\parbox[t]{.4\linewidth}{LL observation}}
  %\State $\bm{\phi}_{tg} \gets$ TG phase
  \State $\{\bm{a}_{res}, \bm{p}_{TG}\} \gets f_{\bm{\theta}_{LL}}(\bm{l}, \bm{s}_{TG}, \bm{o}_{LL})$ \Comment{\parbox[t]{.4\linewidth}{LL execution}}
  \State $\{\bm{a}_{TG}, \bm{s}_{TG}\} \gets$ $f_{TG}(\bm{p}_{TG})$\Comment{\parbox[t]{.4\linewidth}{Trajectory generator execution}}
  %\State $\bm{a} \gets \bm{a}_{TG} + \bm{a}_{NN}$ 
  \State $r \gets$ ExecuteAction($\bm{a}_{res} + \bm{a}_{TG}$)
  \State $d \gets d -1$
  \State $R \gets R + r$
 \EndWhile
\State \textbf{return} $R$
\EndProcedure
\end{algorithmic}
\end{algorithm}

The hierarchical policy structure introduced in \cite{Jain2019HierarchicalRL} that we use for our solution is illustrated in Fig.~\ref{fig:hrlPolicy1}.
The HL is a convolution neural network (CNN) while the LL is a linear fully-connected neural network.
The policy interacts with the robot, which is controlled by combining the output of a trajectory generator (TG) with values computed by the LL's fully connected layer. A TG serves as a parameterized function that computes cyclic leg positions. The LL neural network continuously modulates the TG's phase and amplitude and adjusts the leg trajectories with residuals as needed. More details about the usage of TGs for learning locomotion can be found in~\cite{iscen2018policies}.

Algorithm~\ref{alg:rollout} shows how an episode is executed using a hierarchical policy in which the HL network ($f_{\bm{\theta}_{HL}}$) and LL network ($f_{\bm{\theta}_{LL}}$) have weights $\bm{\theta}_{HL}$ and $\bm{\theta}_{LL}$ respectively. The HL receives task specific exteroceptive observations ($\bm{o}_{HL}$), such as the vision input in our tasks, and issues commands as a latent vector ($\bm{l}$) to LL. HL also decides the duration ($d$) until its next execution. Note that the HL can also optionally receive task-specific inputs (e.g. relative position to goal).
The LL receives proprioceptive observations ($\bm{o}_{LL}$) that include IMU (roll, pitch, roll rate and pitch rate) and motor angles. LL also processes the current latent command ($\bm{l}$) and the current TG state, $\bm{s}_{TG}$.
The LL outputs TG parameters $\bm{p}_{TG}$ and residual motor actuation commands $\bm{a}_{res}$, which are added to TG output $\bm{a}_{TG}$ and executed on the hardware. The environment returns the task reward ($r$) for the robot's action. The HL is invoked again after duration $d$ and the process repeats.

A reinforcement learning problem can be modeled as a Markov Decision Process (MDP) with state space $\mathcal{S}$, action space $\mathcal{A}$, a state transition function $P(\bm{s}_{t+1} | \bm{s}_t, \bm{a}_t)$ and a reward function, $r(\bm{s}_t, \bm{a}_t)$. A policy $\pi_{\Theta}(\bm{s})$, parameterized by a weight vector $\bm{\Theta}$, maps states $\bm{s}$ to actions $\bm{a}$. For a hierarchical policy, $\bm{\Theta}$ is the collection of parameters from all levels ($\bm{\Theta}=\{\bm{\theta}_{HL}, \bm{\theta}_{LL}\}$).
The policy interacts with the MDP for an episode of $T$ timesteps at a time. To jointly learn the parameters $\bm{\theta}_{HL}$ and $\bm{\theta}_{LL}$ of the two levels, we maximize the expected total reward (return) at the end of an episode.

We use an evolutionary algorithm called Augmented Random Search (ARS)~\cite{mania2018simple} to maximize the return. The algorithm proceeds by iteratively estimating gradient of return w.r.t. policy parameters and performing gradient ascent.
% updating them in  the direction of the gradient. 
% The policy is evaluated at a number of points drawn from a normal distribution around the current parameter vector, $\bm{\Theta}$.
% The gradient is estimated as the sum of the top performing parameter perturbations weighted by their return. 

During training, LL skills automatically emerge and are invoked by HL through latent commands ($\bm{l}$) to solve a task.
A trained LL can also be transferred to new tasks in unseen environments. This allows sharing of primitive skills across problems and is faster than learning from scratch on each task. LLs can be transferred by keeping $\bm{\theta}_{LL}$ fixed after training on the original task and re-initializing $\bm{\theta}_{HL}$. Then, during new training only $\bm{\theta}_{HL}$ is updated by ARS.
\begin{figure}[h]
\centering
\begin{subfigure}[t]{.33\columnwidth}
  \centering
  \includegraphics[width=.6\linewidth, height=.6\linewidth, keepaspectratio]{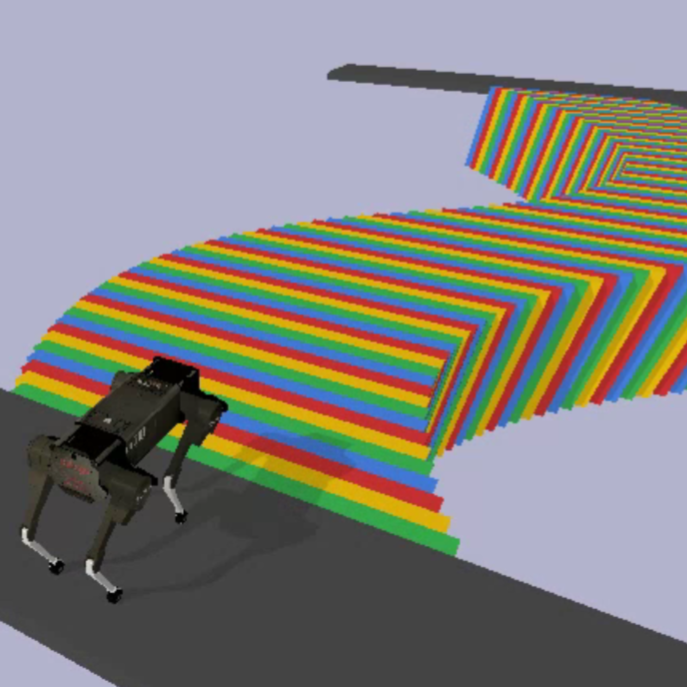}%
  \includegraphics[width=.35\linewidth, height=.35\linewidth, keepaspectratio, frame]{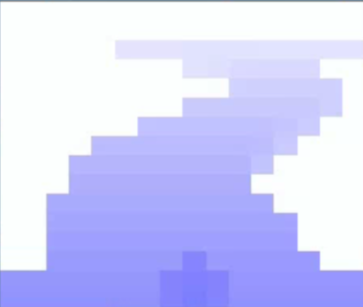}
  \caption{\centering Curved cliff environment.}
  \label{fig:pathenv}
\end{subfigure}%
\begin{subfigure}[t]{.33\columnwidth}
  \centering
  \includegraphics[width=.6\linewidth, height=.6\linewidth, keepaspectratio]{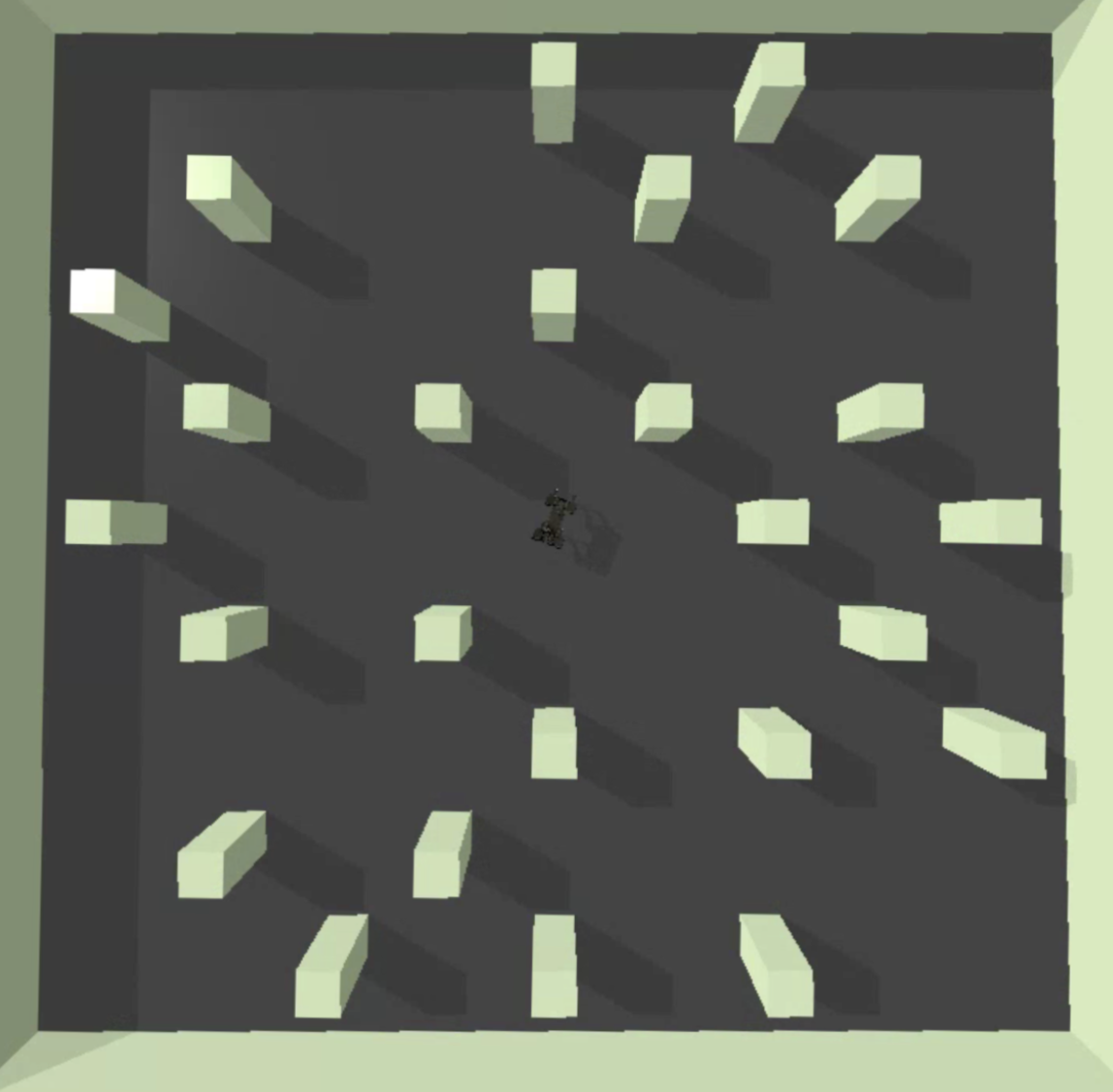}%
  \includegraphics[width=.35\linewidth, height=.35\linewidth, keepaspectratio, frame]{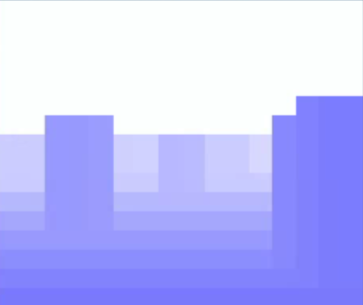}
  \caption{\centering Maze traversal task.}
  \label{fig:navexenv}
\end{subfigure}%
\begin{subfigure}[t]{.33\columnwidth}
  \centering
  \includegraphics[width=.6\linewidth, height=.6\linewidth, keepaspectratio]{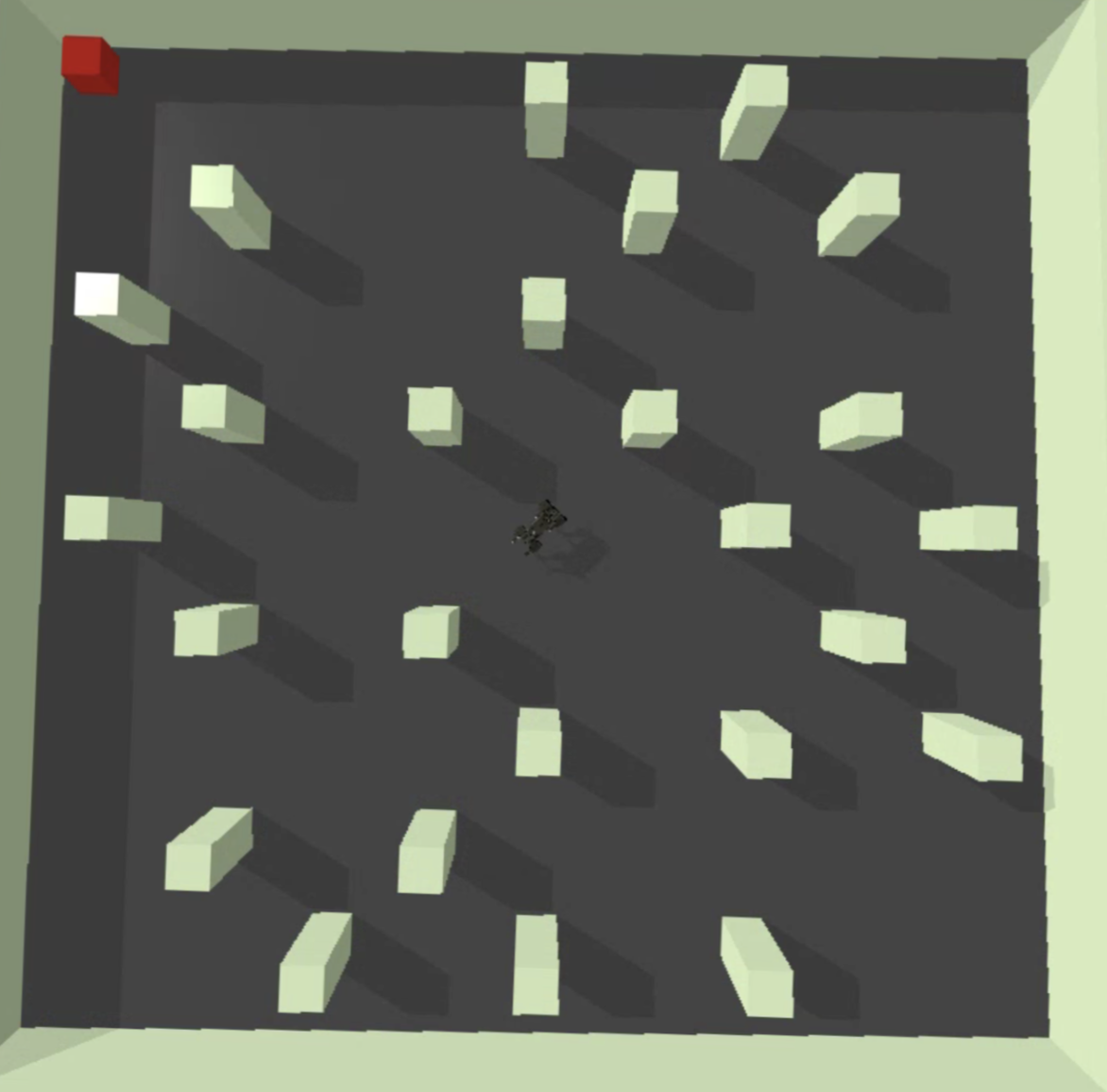}%
  \includegraphics[width=.35\linewidth, height=.35\linewidth, keepaspectratio, frame]{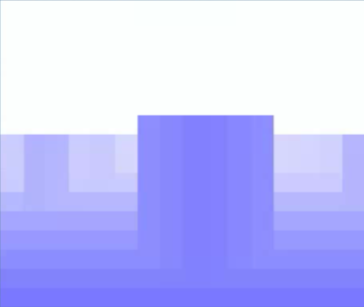}
  \caption{\centering Goal finding task in maze.}
  \label{fig:navmixenv}
\end{subfigure}
\caption{\centering Task environments and vision inputs from depth camera.} % \atilsays{camera outputs or input?}}
\label{fig:envs}
\end{figure}

\section{Experimental Setup}
\label{sec:task_details}
We use the Laikago quadruped robot from Unitree\footnote{\href{http://www.unitree.cc/e/action/ShowInfo.php?classid=6&id=1}{unitree.cc}}. This robot is $\SI{60}{\cm}$ tall, has $12$ degrees of freedom ($3$ per leg) and weighs about $\SI{22}{\kg}$. The swing and extension of each leg is controlled by a PD position controller provided with the robot. 
We train our policies in simulation using PyBullet~\cite{pybulletcoumans,tan2018sim}.
Our tasks are set up in two environments: a curved cliff and a maze.

\textbf{Curved Cliff Environment.} In this environment, the robot starts from the origin and a curved cliff lies ahead of it. The robot can observe the environment through a first-person depth-camera view, angled down slightly to see the cliff. Fig.~\ref{fig:pathenv} shows a still from this environment and a sample camera input. The robot's task in this environment is to progress forward as fast as possible. To accomplish this, it needs to learn to steer in order to follow the curves of the cliff and avoid falling off the edge. The shape of the cliff curve is randomized for each episode. The reward function is specified as the capped ($v_{cap}$) velocity of the robot along the x direction:

\begin{align}
f_{v_{cap}}(r) &= \max(-v_{cap},\min(r, v_{cap}))\\
    r_{cc}(t) &= f_{v_{cap}}(x(t) - x(t-1)).
    % R_{cc} &= \sum_{t \ge 1}r_{cc}(t),
\end{align}

\textbf{Maze Environment.} For the maze environment, the robot is placed in the middle of a walled $13 \times \SI{13}{\meter\squared}$ arena uniformly filled with pillars. An episode starts with randomly orientated robot observing the world with a depth-camera looking straight ahead. This environment and a sample camera image is shown in Fig.~\ref{fig:navexenv}. We set up two tasks in this environment: maze traversal and goal finding. 
For the maze traversal task, the robot needs to keep going further away from its starting point (origin). The optimal behavior constitutes stable forward walking and steering to avoid colliding with pillars and boundary walls. For this task, the robot also observes its position and orientation relative to the origin, $\mathbf{x}$. The reward function for this task is as follows:

\begin{align}
    %r_{mt}(t) &= f_{v_{cap}}(\dist(\mathbf{x}(t)) - \dist(\mathbf{x}(t-1))
    r_{mt}(t) &= f_{v_{cap}}(\left\|\mathbf{x}(t) \right\|-\left\|\mathbf{x}(t-1) \right\|).
    % R_{mt} &= \sum_{t \ge 1}r_{mt}(t),
\end{align}

In the goal finding task, the robot needs to reach a goal randomly placed in one of the $4$ corners of the maze for each episode (see Fig.~\ref{fig:navmixenv}). Along with the camera input, it observes its position and orientation relative to the goal. To successfully find the goal it needs to learn to align itself in the direction of the goal along with all the skills for maze traversal. The reward function is given by:

\begin{align}
    % v(t) &= \dist(\mathbf{x} (t-1) - \mathbf{g}) - \dist(\mathbf{x} (t) - \mathbf{g}) \\
    r_{gf}(t) &= f_{v_{cap}}(\left\|\mathbf{x} (t-1) - \mathbf{g}\right\| - \left\|\mathbf{x} (t) - \mathbf{g}\right\|)\\
    % \omega &= \dist(\mathbf{x} (t)) / \dist(\mathbf{g} (t)) \\
    r(t) &= \omega r_{gf}(t) + (1 - \omega) r_{mt}(t) ; \quad \omega = \left\|\mathbf{x} (t)\right\| / \left\|\mathbf{g}\right\|,
    % R_{gf} &= \sum_{t \ge 1}r(t),
\end{align}

where $\mathbf{g}$ is the position of the goal. The reward is a weighted average of the maze traversal reward term $r_{mt}$, and a term for progression towards the goal position, $r_{gf}$, based on $\omega$. The variable $\omega$ corresponds to the fraction of the distance travelled relative to the total distance from the goal to the origin. The reward is dominated by $r_{mt}$ when the robot is close to the origin and becomes more defined by $r_{gf}$ as it gets closer to the goal. This reward function encourages the robot to learn locomotion skills in the early stages of training. Without $r_{mt}$, the robot doesn't experience any positive reinforcement for stable walking unless it happens to walk in the goal direction.

In all tasks, the episode terminates if the robot loses its balance, falls off a cliff, collides with a pillar or boundary wall, or if the episode reaches $6000$ LL robot control time steps ($\SI{12}{\s}$). Additionally, in the goal finding task, the episodes terminates when the robot comes within $\SI{0.5}{\meter}$ of the goal position. We set $v_{cap}$ to $\SI{0.002}{\meter}$ in all experiments, which corresponds to $\SI{1}{\meter\per\second}$.

\textbf{Solution Implementation details.} The high level contains a CNN that receives a $16 \times 16 \times 1$ depth camera input. It has $3$ convolutional layers of $3 \times 3$ filters with output channels $4, 8$, and $8$, followed by a pooling layer with filter of size $2 \times 2$ applied with a stride of $2$. Output from the pooling layer is flattened and transformed into a $10D$ feature vector through a fully-connected layer with $\tanh$ activation. If present, the task-specific  HL inputs (relative position in the maze environment) are concatenated with the feature vector. It is then fed into a fully-connected layer to produce an output clipped between $-1$ and $1$. For most of our experiments we use a $3D$ output, with the first dimension corresponding to the HL duration ($d$) and the rest to the latent command ($\bm{l}$). The duration is calculated by linearly scaling the output to a value between $50$ - $300$ time-steps ($\approx \SI{1.5}{\hertz} - \SI{10}{\hertz}$). The latent command concatenated with IMU, motor angles, and  trajectory generator (TG) state is fed to LL linear fully-connected network to output the residual motor commands and PMTG parameters. HL network has around $3000$ parameters and LL has around $300$. For comparison, we also train a non-hierarchical CNN with same convolutional and pooling layers as above. The feature vector is concatenated with all other sensor observations and fed to $2$ fully connected layers (hidden layer size is $10$) before producing the actions.

The trajectory generator (TG) is based on the \emph{Policies Modulating Trajectory Generators} (PMTG) architecture,  which has shown success at learning diverse primitive behaviors for quadruped robots~\cite{iscen2018policies}. 
As mentioned above, LL observes the PMTG state ($\bm{s}_{TG}$), which specifies the position along a periodic leg trajectory, and updates the PMTG parameters at every time-step.

We train the policies using a distributed ARS implementation. For each optimization iteration, we evaluate policy perturbations on $64$ parallel workers. Since all our tasks are randomized, we take average of return from $3$ environment episodes to evaluate each perturbation. The number of perturbations evaluated, gradient step size, number of top perturbations used for gradient estimation, and the standard deviation for generating new perturbations are all determined by hyper-parameter tuning using a gaussian process bandits approach~\cite{Vizier}.

\begin{figure}[h]
\begin{subfigure}[c]{0.30\columnwidth}
\centering
  \includegraphics[width=.9\linewidth]{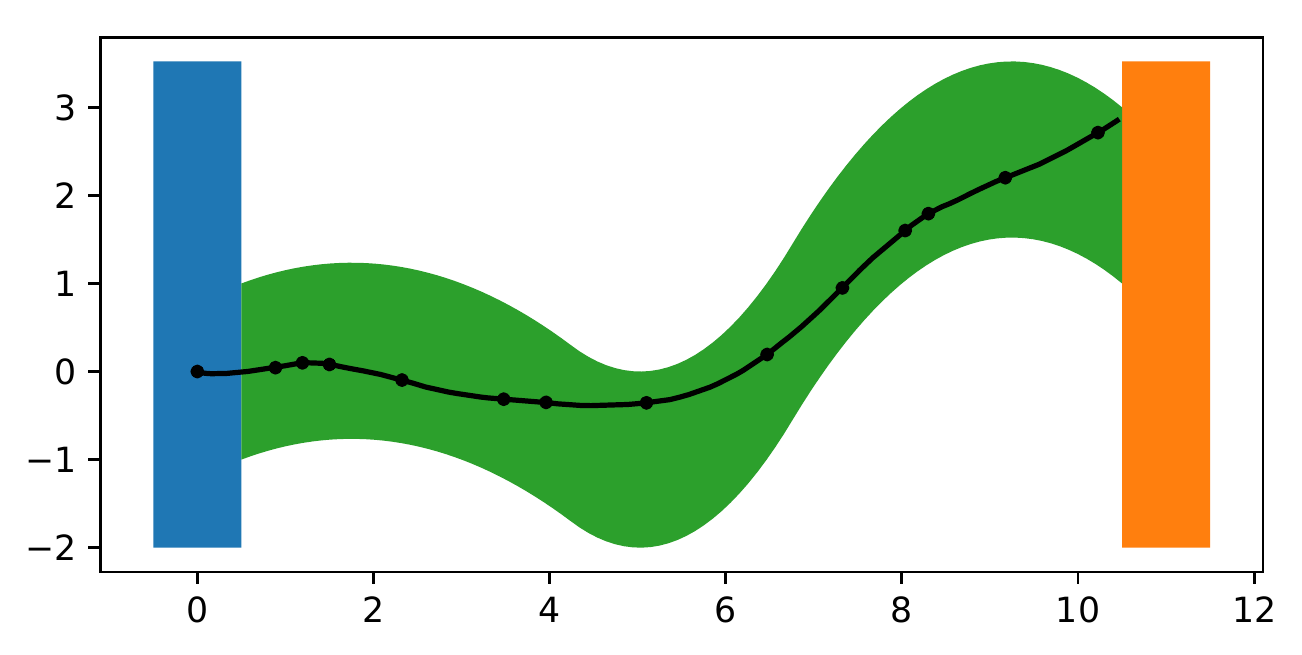} \\
  \includegraphics[width=.9\linewidth]{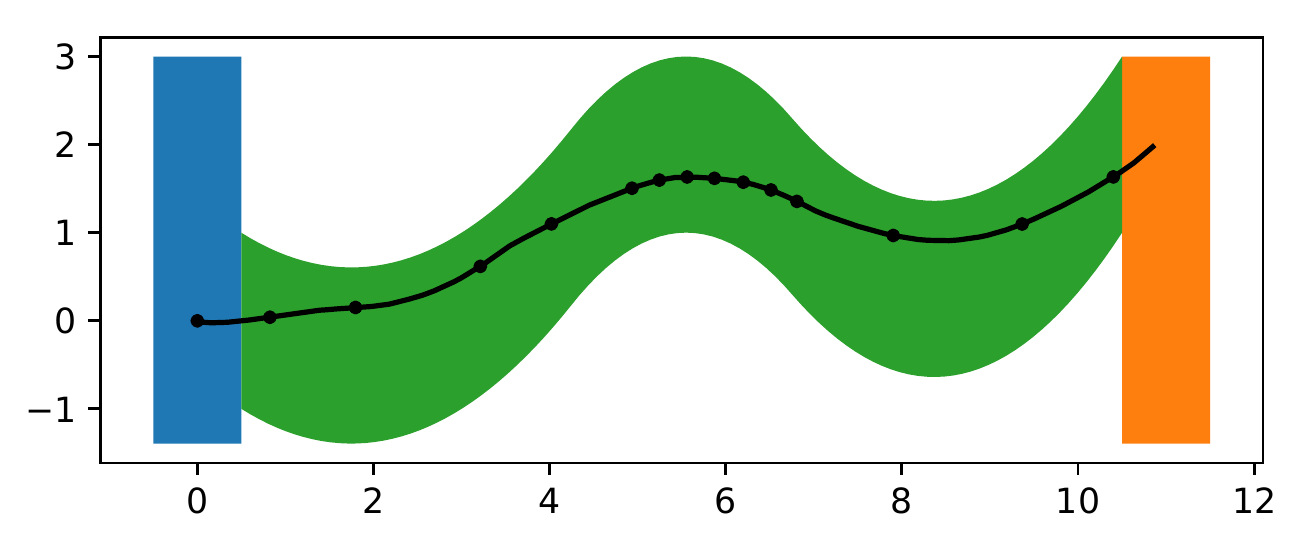} \\
  \includegraphics[width=.9\linewidth]{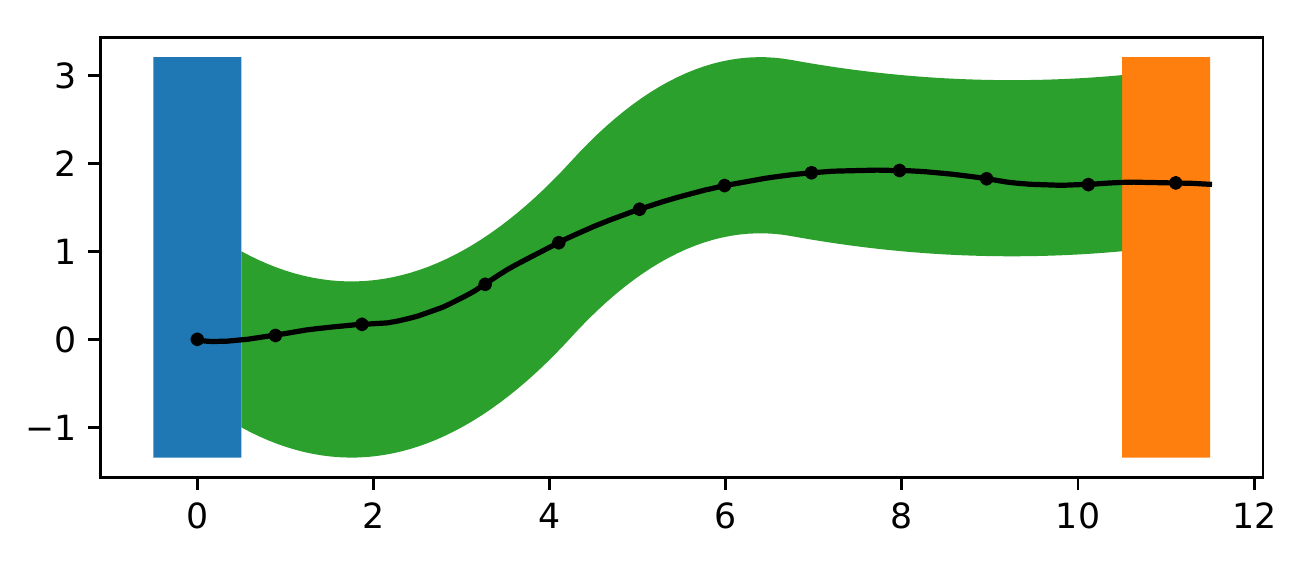}
\caption{Curved cliff task.}
\label{fig:cliffTrajectories}
\end{subfigure}%
\begin{subfigure}[c]{0.35\columnwidth}
\centering
  \includegraphics[width=.65\linewidth,height=.63\linewidth]{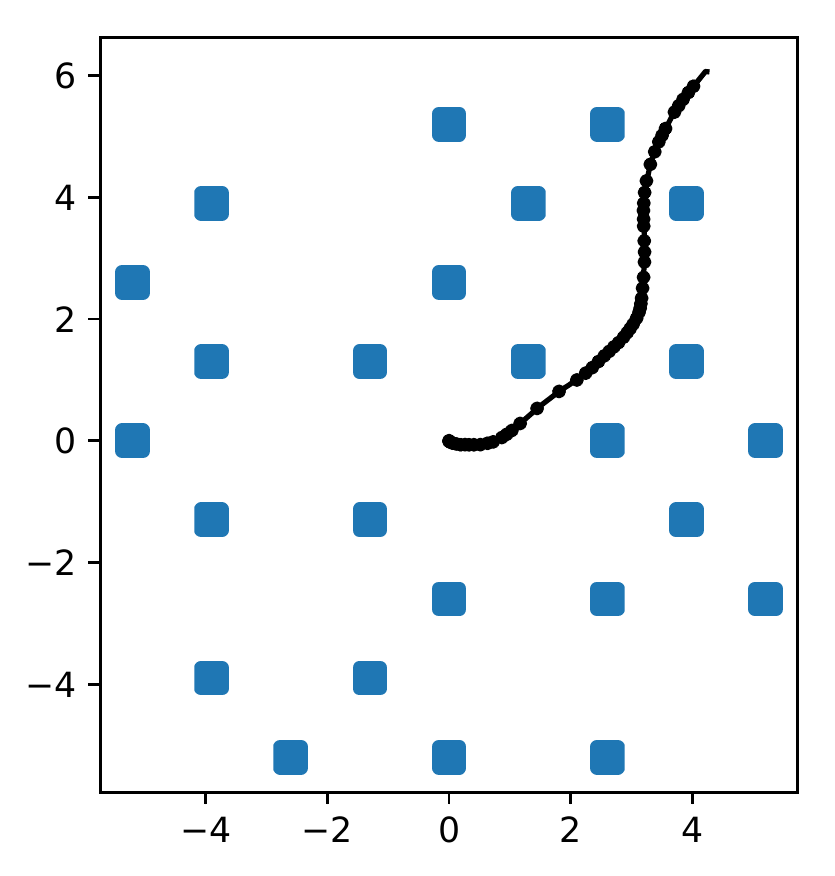} \\
  \hspace{3pt}\includegraphics[trim=50 100 150 100,clip,width=.37\linewidth,keepaspectratio,frame]{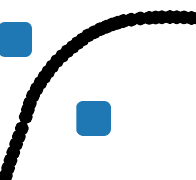}%
  \includegraphics[trim=130 190 73 27.7,clip,width=.37\linewidth,keepaspectratio,frame]{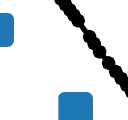}
\caption{Maze traversal task.}
\label{fig:mazeTraversalTrajectories}
\end{subfigure}%
\begin{subfigure}[c]{0.35\columnwidth}
\centering
  \includegraphics[width=.67\linewidth]{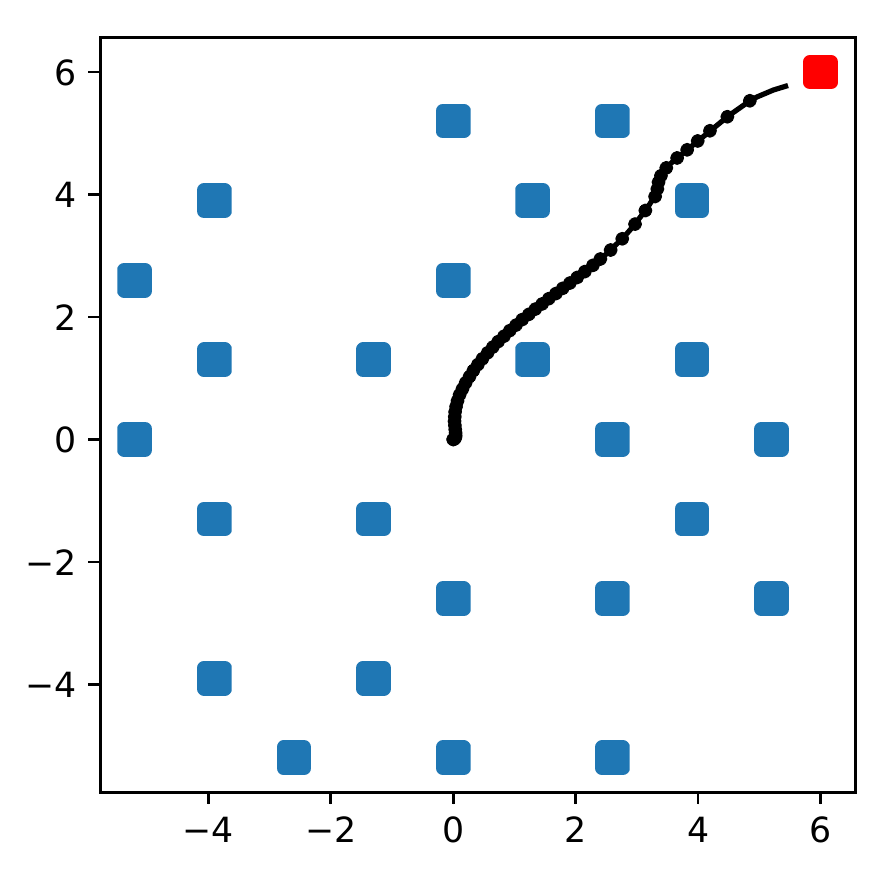} \\
  \includegraphics[trim=185 25 15 175,clip,width=.37\linewidth,keepaspectratio,frame]{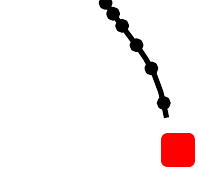}%
  \includegraphics[trim=100 100 100 100,clip,width=.37\linewidth,keepaspectratio,frame]{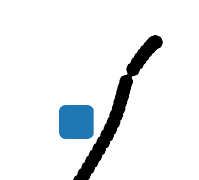}
\caption{Goal Finding Task.}
\label{fig:goalFindingTrajectories}
\end{subfigure}
\caption{Robot trajectories in simulated environments. Dots indicate HL execution ($d$). Axes in $\si{\meter}$.}
\label{fig:simTrajectories}
\end{figure}

\section{Experimental Results}
Our hierarchical policies are able to complete all $3$ visual navigation tasks described in Sec.~\ref{sec:task_details} by learning locomotion directly from vision input. Fig.~\ref{fig:simTrajectories} shows the trajectories of the trained robot in simulation for the $3$ tasks.
Dot markers, along the trajectories, show the points at which HL becomes active and computes the next latent command ($\bm{l}$) and duration ($d$). Notice that for solving the curved cliff task (Fig.~\ref{fig:cliffTrajectories}), the HL takes decisions more frequently (small $d$) when sharp turns are made to avoid falling off the cliff. In straighter regions, HL executions are sparser (large $d$). 
For the goal finding task, HL takes sparser decisions when the goal is close (Fig.~\ref{fig:goalFindingTrajectories}). The robot efficiently turns in-place to face the goal using dynamic leg movements which are difficult to hand-design and tune. 

% \subsection{Comparison with Baseline Neural Network Policy}
\begin{figure*}
\centering
\begin{subfigure}[t]{.33\columnwidth}
  \centering
  \includegraphics[width=.95\linewidth]{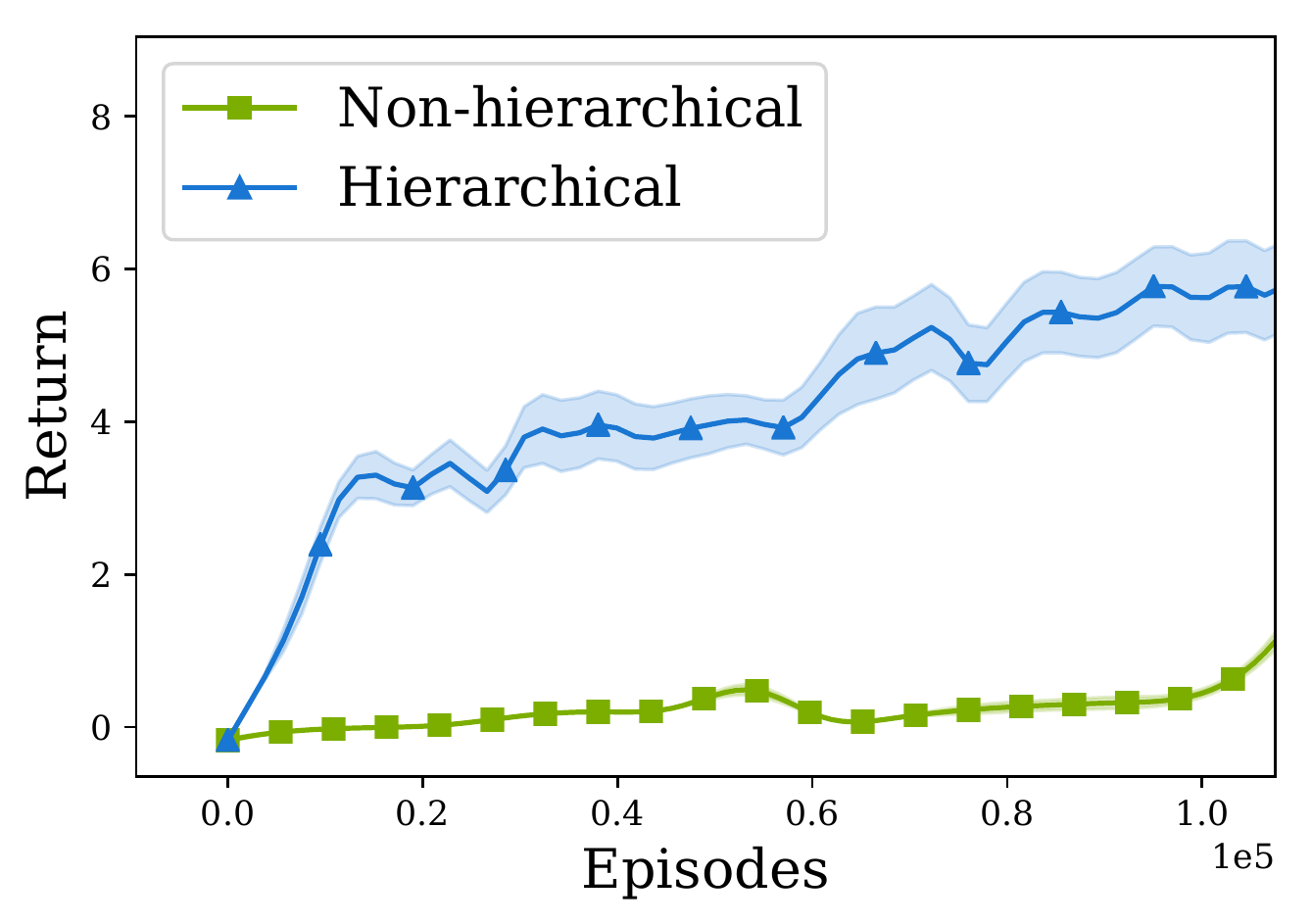}
  \caption*{\centering Curved cliff task.}
  \label{fig:hrlvsnn_path}
\end{subfigure}%
\begin{subfigure}[t]{.33\columnwidth}
  \centering
  \includegraphics[width=.95\linewidth]{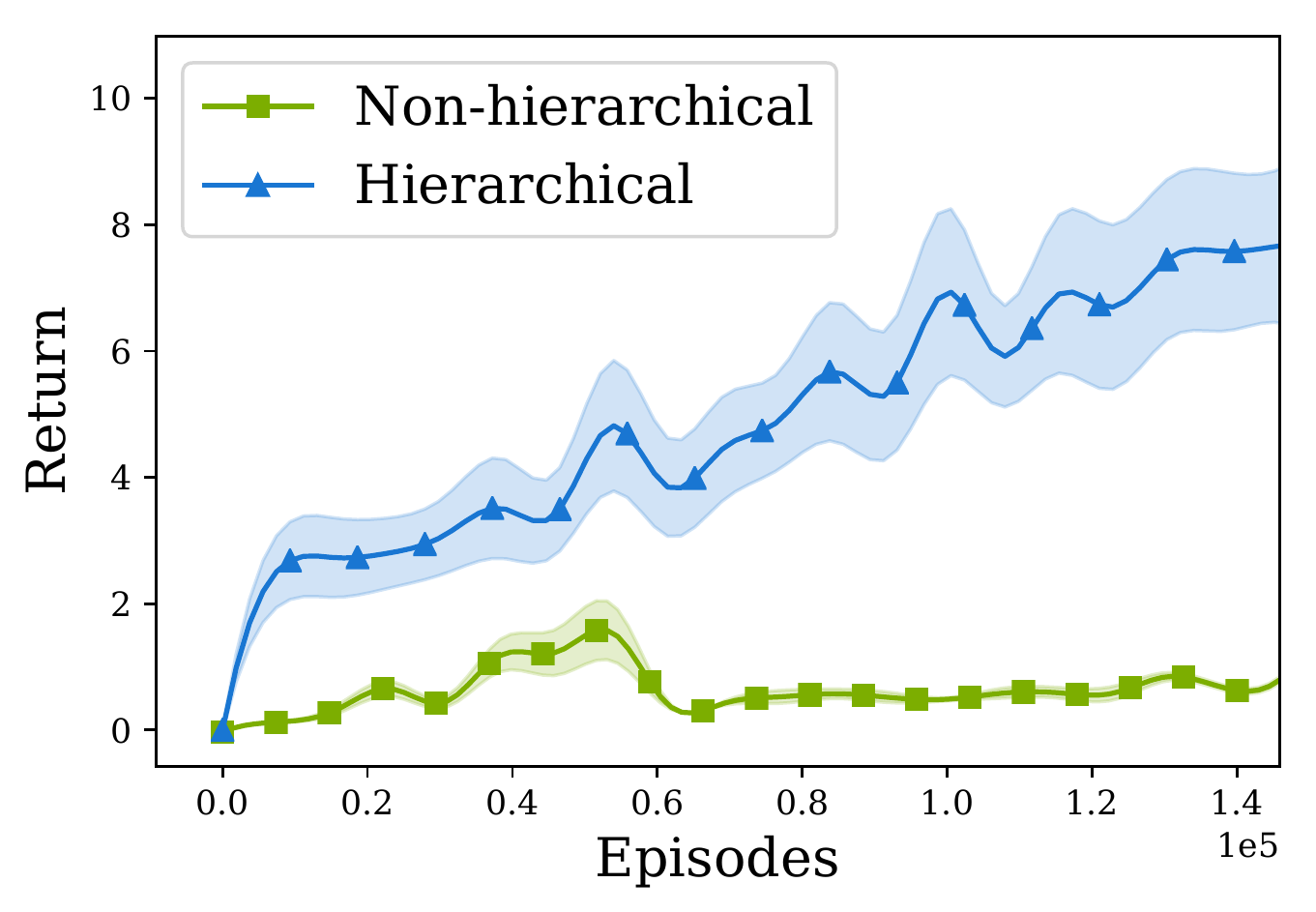}
  \caption*{\centering Maze traversal task.}
  \label{fig:hrlvsnn_navex}
\end{subfigure}%
\begin{subfigure}[t]{.33\columnwidth}
  \centering
  \includegraphics[width=.95\linewidth]{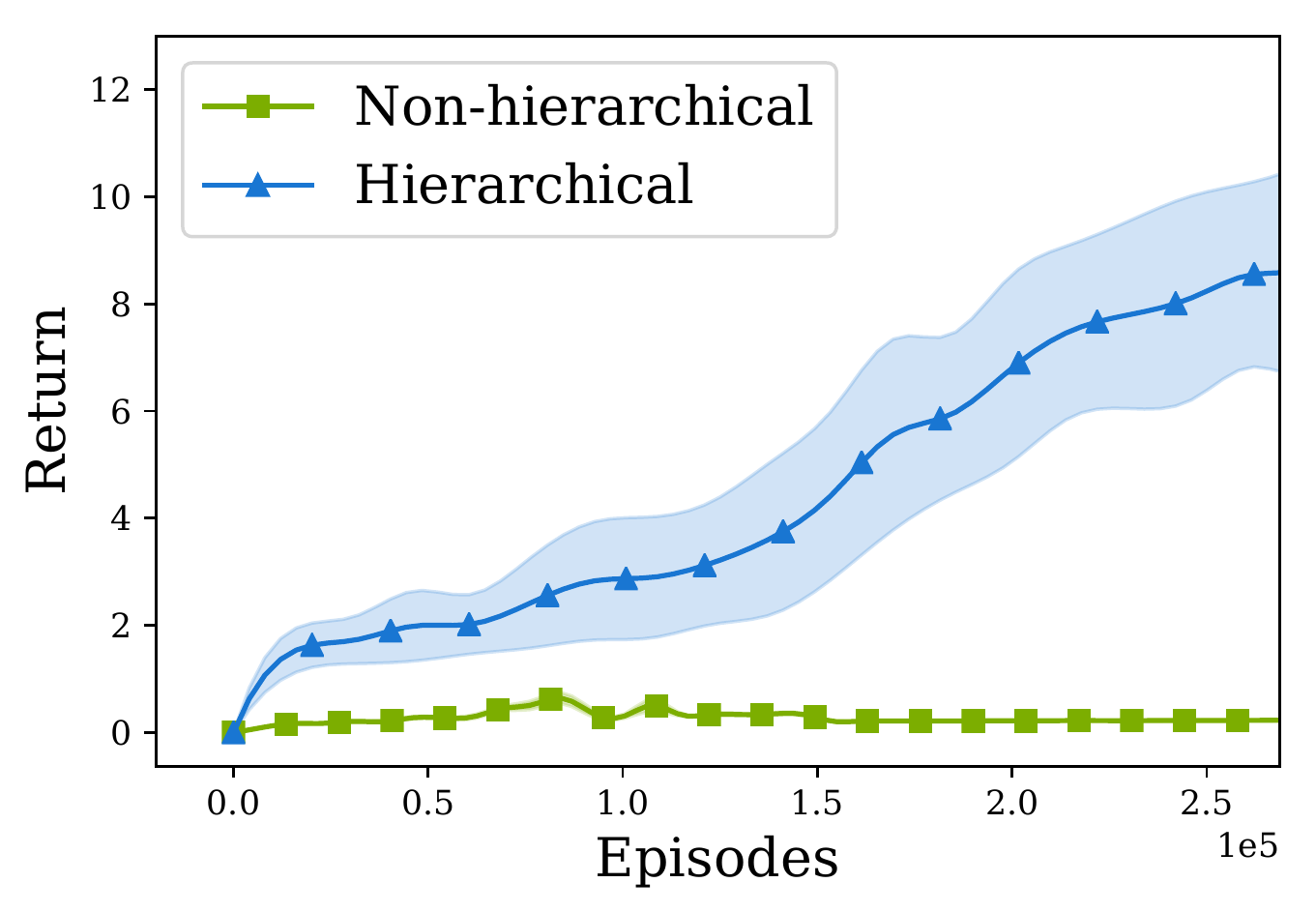}
  \caption*{\centering Goal finding task.}
  \label{fig:hrlvsnn_navmix}
\end{subfigure}
\caption{Learning curves for HRL and CNN policies.}
\label{fig:hrlvsnn}
% \vspace{1cm}
\end{figure*}

We compare the learning curves for our hierarchical policies described in Sec.~\ref{sec:method} with non-hierarchical CNN policies on these $3$ tasks (refer to Fig. \ref{fig:hrlvsnn}). The plot lines show the average return across $\approx 450$ environment episodes. The shaded region denotes the standard deviation. The return is plotted against the total training episodes. We see that in all $3$ cases our method largely outperforms the baseline, completing each task at the end of the training.

\begin{figure}[h]
\centering
\hspace{-0.1in}
\begin{minipage}[t]{.35\textwidth}
\centering
  \includegraphics[width=.95\linewidth]{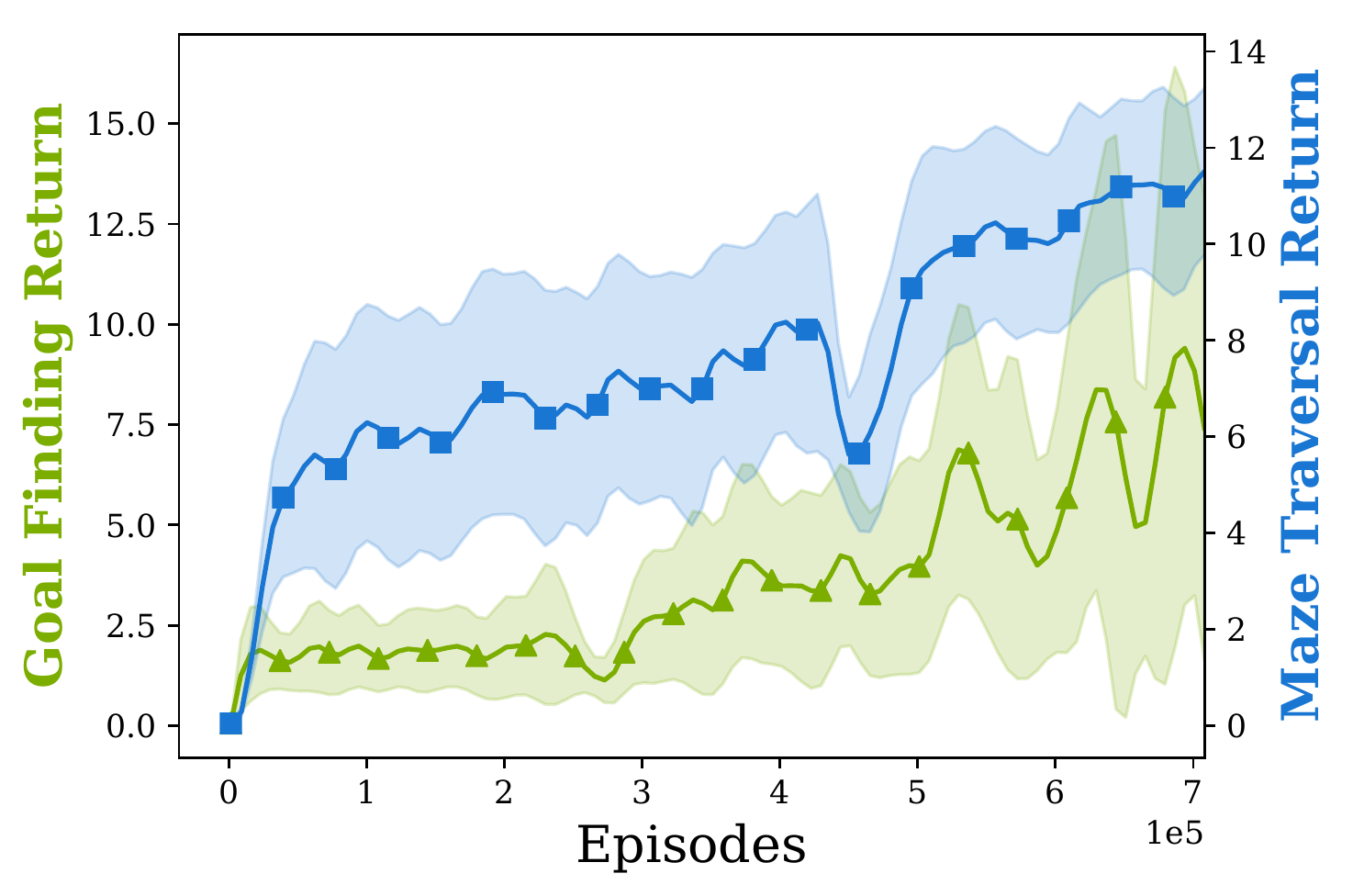}
\captionof{figure}{\centering Learning HRL policies with $10^5$ parameters.}
\label{fig:big_cnn}
\end{minipage}%
\begin{minipage}[t]{.66\textwidth}
\centering
\begin{minipage}[t]{.5\columnwidth}
  \centering
  \includegraphics[width=.97\linewidth]{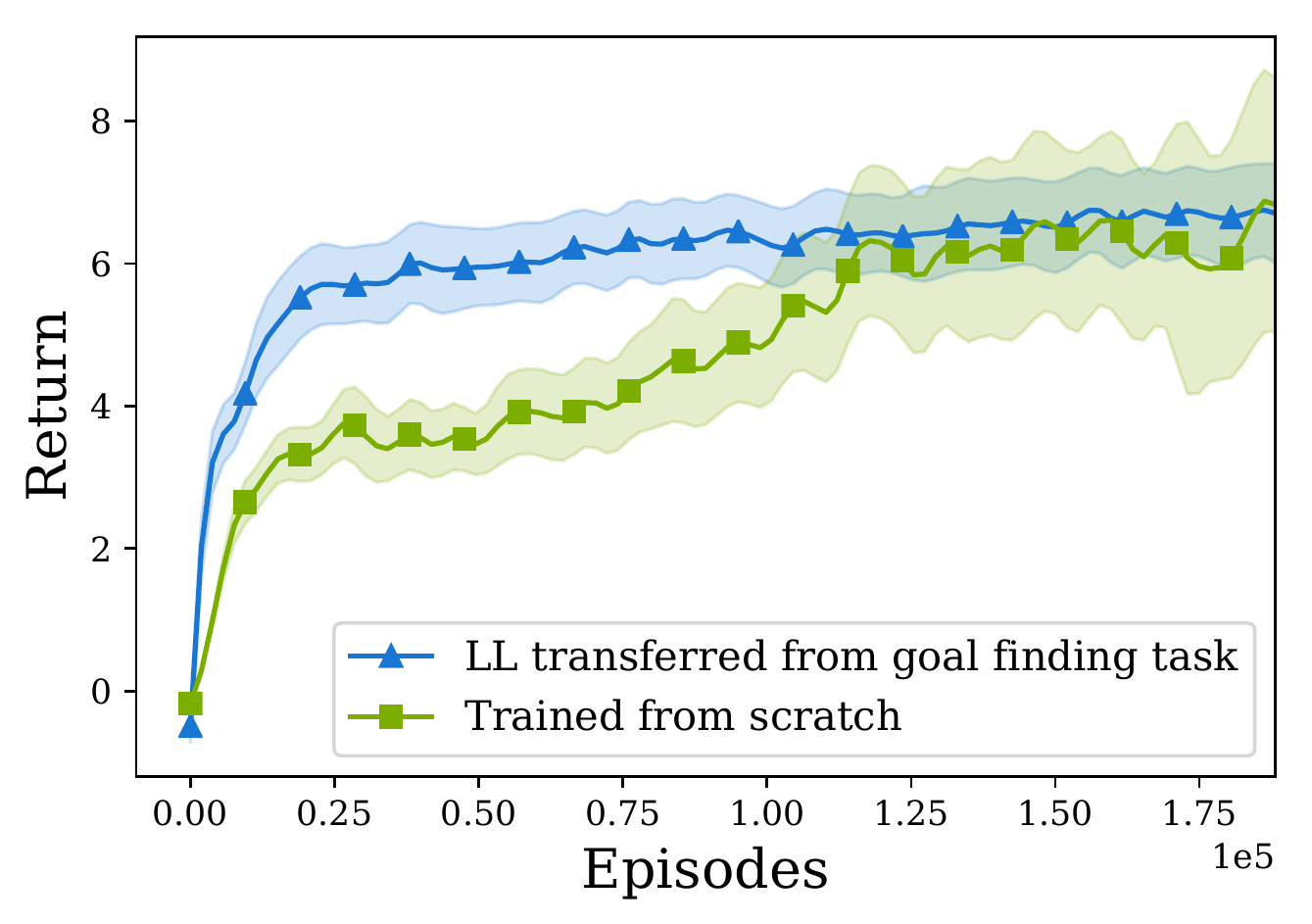}
  \caption*{\centering Curved cliff.}
\end{minipage}%
\begin{minipage}[t]{.5\columnwidth}
  \centering
  \includegraphics[width=.97\linewidth]{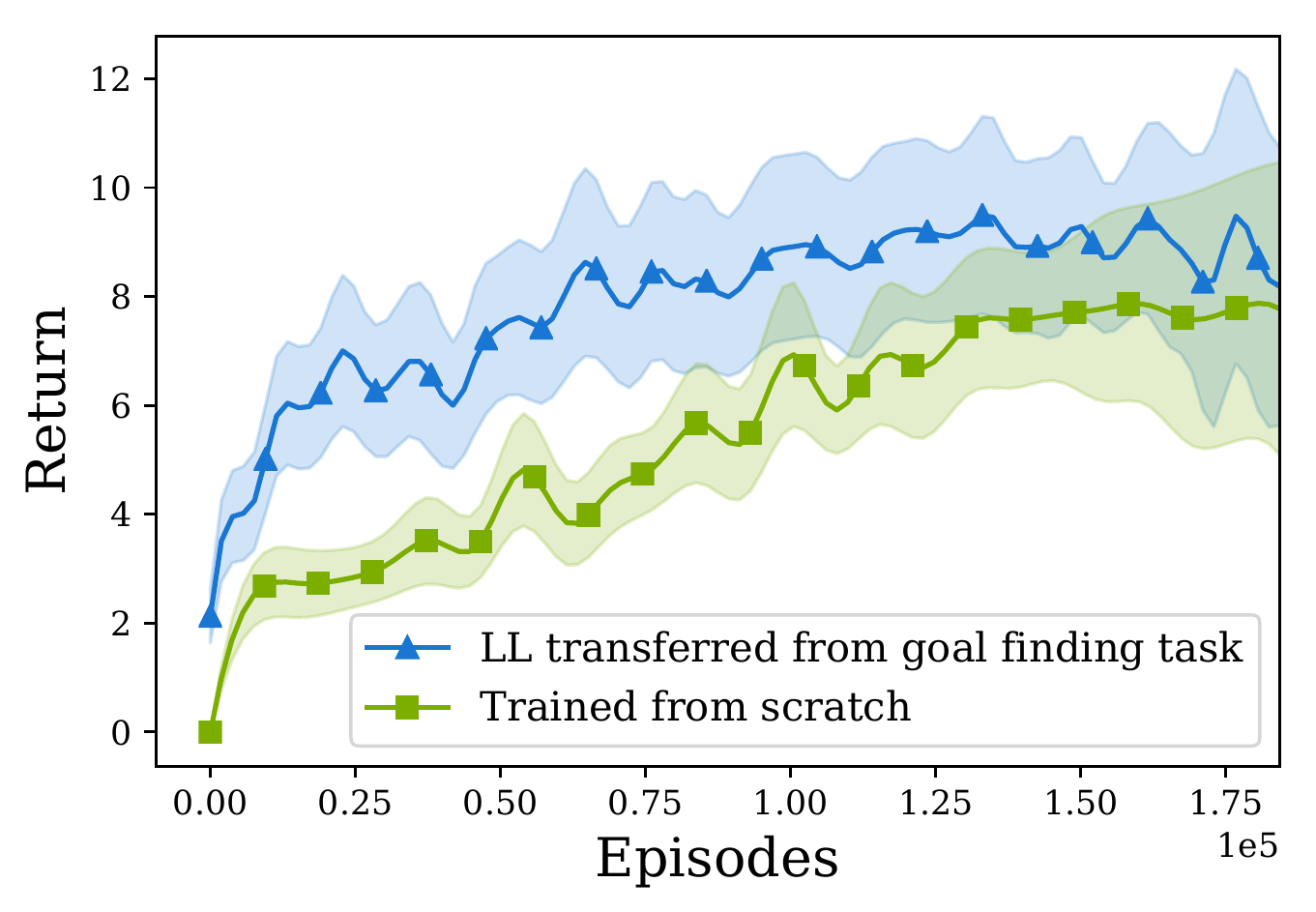}
  \caption*{\centering Maze traversal.}
\end{minipage}
\captionof{figure}{\centering Transferring LL from goal finding task.}% \atilsays{We can change these x axis to be divided by 1e4.}}
\label{fig:transfer}
\end{minipage}
\end{figure}

% \pagedepth\maxdimen
% \begin{wrapfigure}[22]{R}{0.4\linewidth}
% \centering
% \begin{subfigure}{\linewidth}
%   \centering
%   \includegraphics[width=.95\linewidth]{figures/bigCNNs/big_navex.pdf}
%   \caption*{\centering Maze traversal task.}
% \end{subfigure} \\
% \begin{subfigure}{\linewidth}
%   \centering
%   \includegraphics[width=.95\linewidth]{figures/bigCNNs/big_navmix.pdf}
%   \caption*{\centering Goal finding task.}
% \end{subfigure}
% \caption{Training curves for hierarchical policies with $130K$ parameters.}
% \label{fig:big_cnn}
% \end{wrapfigure}
% \paragraph{Larger CNN Policy to Handle Higher Resolution Vision Input.} 
Though we trained policies with approximately $10^3$ weights (Sec.~\ref{sec:task_details} for details) in most of our experiments, we show that larger CNNs with over $10^5$ weights can be trained using ARS. Fig.~\ref{fig:big_cnn} shows the learning curve for the maze traversal and goal finding tasks. The image resolution is $32 \times 32$ and we use $2$ additional $32D$ fully connected layers at the end of the CNN. 
% ARS successfully optimized the weights of our policies and achieved optimal performance on these tasks. 

\paragraph{Transferring Low Level Policies between Tasks.} LLs from learned policies can be reused for new tasks and environments as shown in Fig.~\ref{fig:transfer}. We first trained a policy with $2D$ latent commands for the goal finding task. The LL only has access to proprioceptive sensors, which forces it to learn generic steering and turning-in-place primitives. We reuse these skills for training a new policy in the maze traversal and curved cliff tasks based on the goal finding LL. LL weights are initialized from a pre-trained policy and frozen during the new HL policy training. In both of these cases we observe that this improves the learning efficiency. 
As can be expected, LL policies trained on the simpler and more restricted cliff task did not yield good performance when transferred to the maze traversal or goal finding tasks.
In future work, we plan to explore fine-tuning transferred LL weights so that it can adapt to new tasks if previously learned skills do not suffice.

\section{Analysis of Hierarchical Policies}
\begin{figure}[h]
\centering
\hspace{-0.1in}
\begin{minipage}[t]{.64\linewidth}
\centering
\begin{minipage}[t]{.5\linewidth}
  \centering
  \includegraphics[width=.95\linewidth]{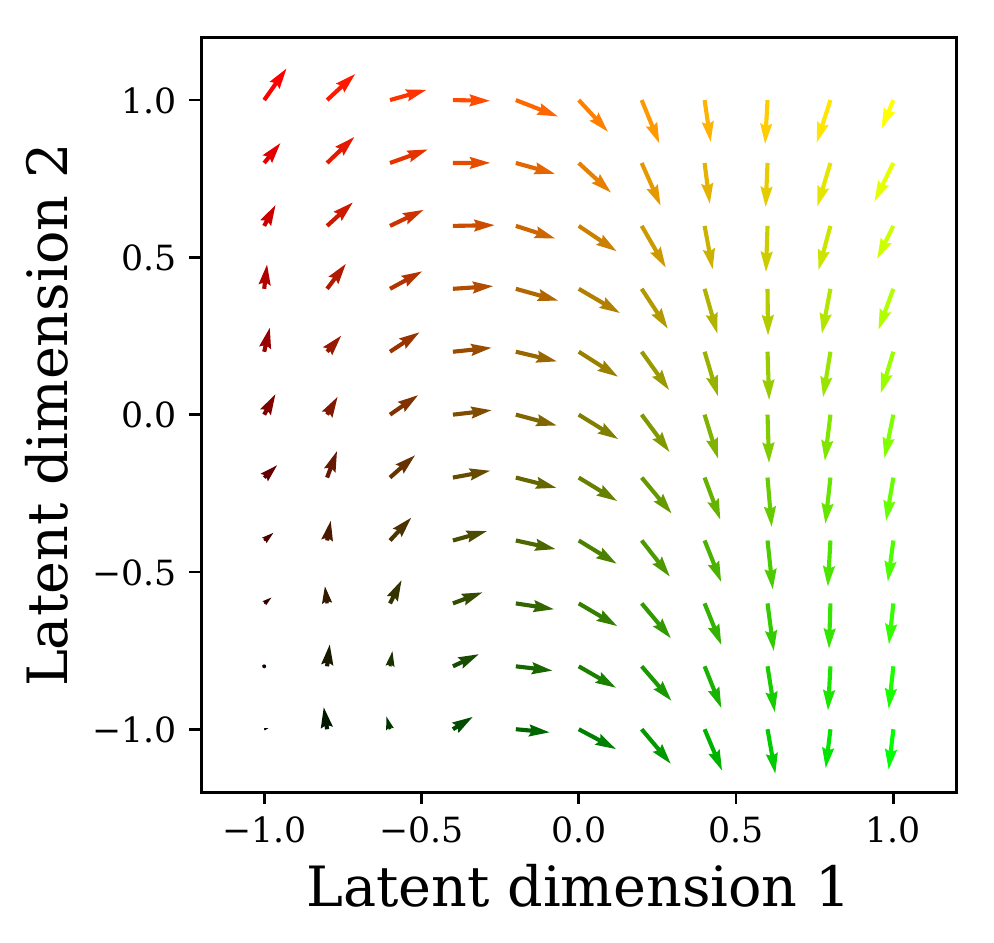}
  \caption*{\centering 2D latent command space.}
\end{minipage}%
\begin{minipage}[t]{.5\linewidth}
  \centering
  \includegraphics[width=.86\linewidth]{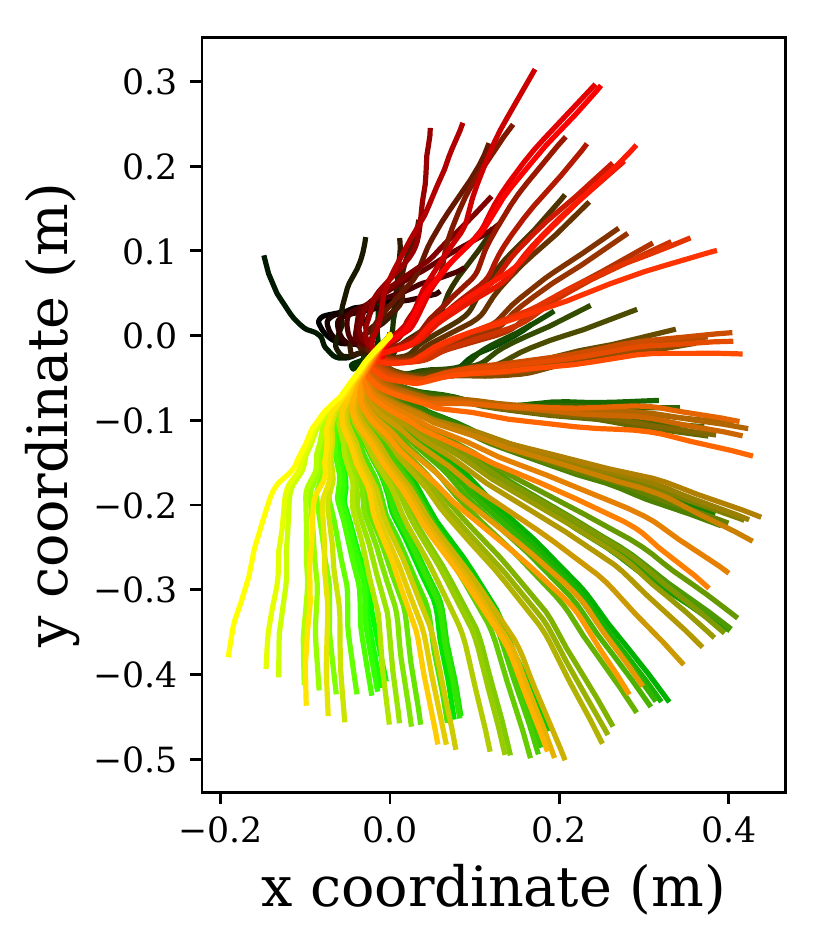}
  \caption*{\centering LL trajectories.}
\end{minipage}
\captionof{figure}{\centering Latent command space analysis.}
\label{fig:latentspaceanalysis}
% \vspace{1cm}
\end{minipage}\hfill
\begin{minipage}{.35\linewidth}
    \vspace{-1.3in}
    \centering
    \includegraphics[width=.95\linewidth]{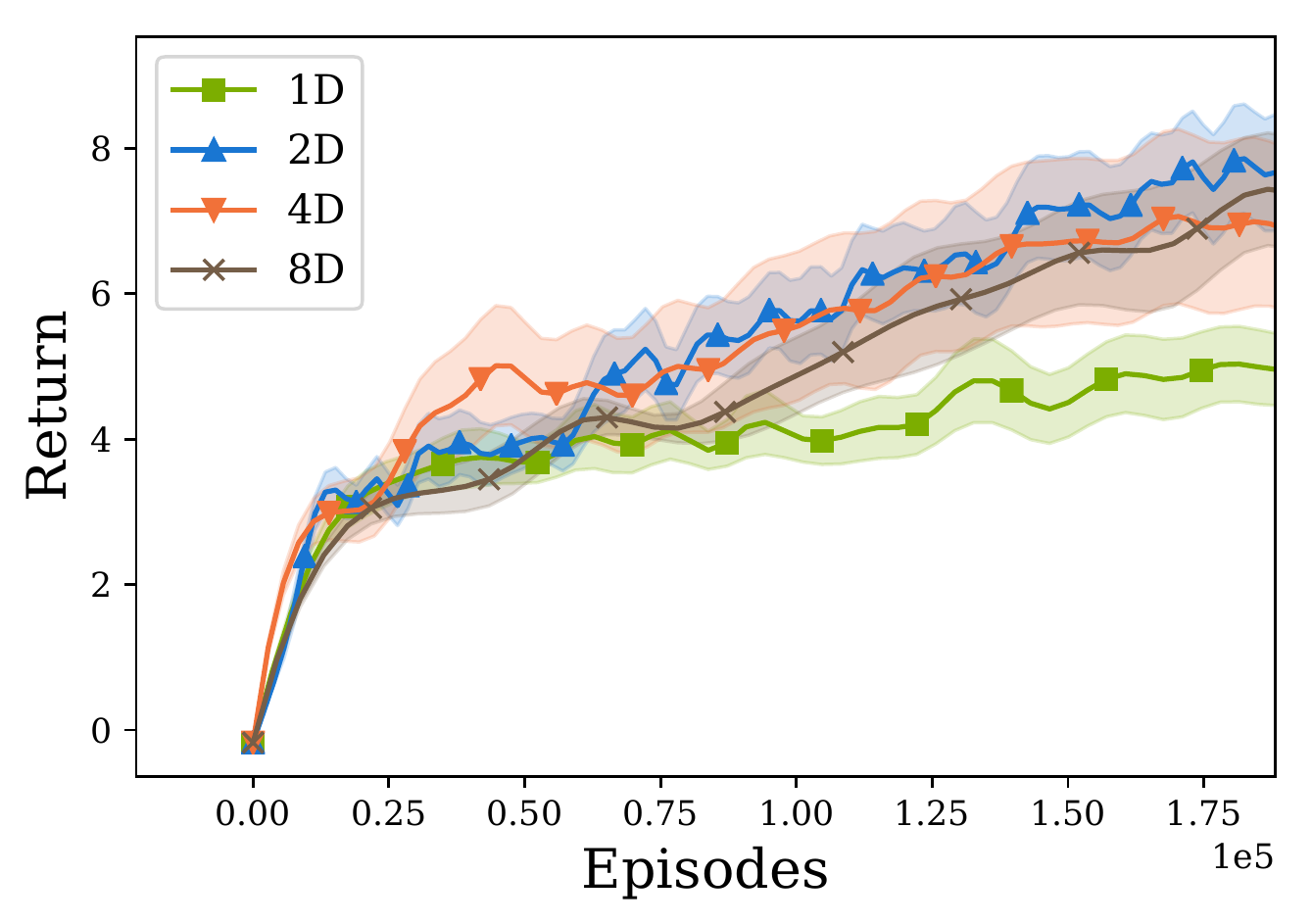}
    \captionof{figure}{\centering Latent command dimensions comparison.}
    \label{fig:lcdim_compare}
\end{minipage}
\end{figure}
\textbf{Analysing 2D Latent Command Space.} To better understand the behavior of the learned policy, we visualize a $2D$ latent command space for the goal finding task (Fig.~\ref{fig:latentspaceanalysis}). For a learned hierarchical policy, we evaluated the low level with artificial latent command inputs taken from a uniform grid over the whole command space. Those behaviors are shown on the right of the figure in the form of robot trajectories running for $\SI{1}{\second}$ in the XY coordinate plane. On the left we show the latent command space points for which these behaviors emerged in corresponding colors. For each point, we generate a vector summarizing the LL trajectory. Vector direction shows the movement direction of the whole LL trajectory and the length is proportional to the distance covered. This visualization shows how the latent space is used to smoothly transition between robot steering behaviors of varying velocities that have emerged automatically.

% On the left in Fig.~\ref{fig:latentspaceanalysis} we show the movement direction of the robot when giving different points in latent space as commands to the LL and executing the LL for $\SI{1}{\second}$. The length of the arrow is proportional to the distance covered. Corresponding color-coded robot trajectories are shown on the right. We can observe that robot steering behaviors of varying velocities emerge automatically as LL behaviors. The HL uses these behaviors to navigate different parts of the environment. 
% We can also observe the difference in emerged low level behaviors between these tasks. According to the needs of the environment, the low level policy for the curved cliff task learns to cover more distance with small steering angles, while that for the goal finding task learns more aggressive steering and turning in place required for aligning itself in goal direction and obstacle avoidance.

\textbf{Influence of Latent Command Dimension.} The comparison between hierarchical policies with $1, 2, 4$, and $8$ dimensional latent command space (LCS) is shown in Fig.~\ref{fig:lcdim_compare}. These policies are trained on the curved cliff task. It is clear that the $1D$ LCS is too restrictive for this task and the policy is not able to achieve optimal performance. The $2, 4$, and $8D$ LCS perform similarly. It is promising to see $2D$ LCS reaching optimal performance, since low dimensional LCS has many benefits: they can be easily visualized and interpreted, they are easy to control and hence amenable to transfer, and they reduce the network size making it easier to train.

\begin{figure}
% \centering
\begin{subfigure}[c]{.4\columnwidth}
\hspace{-20pt}
%   \centering
  \includegraphics[width=.95\linewidth]{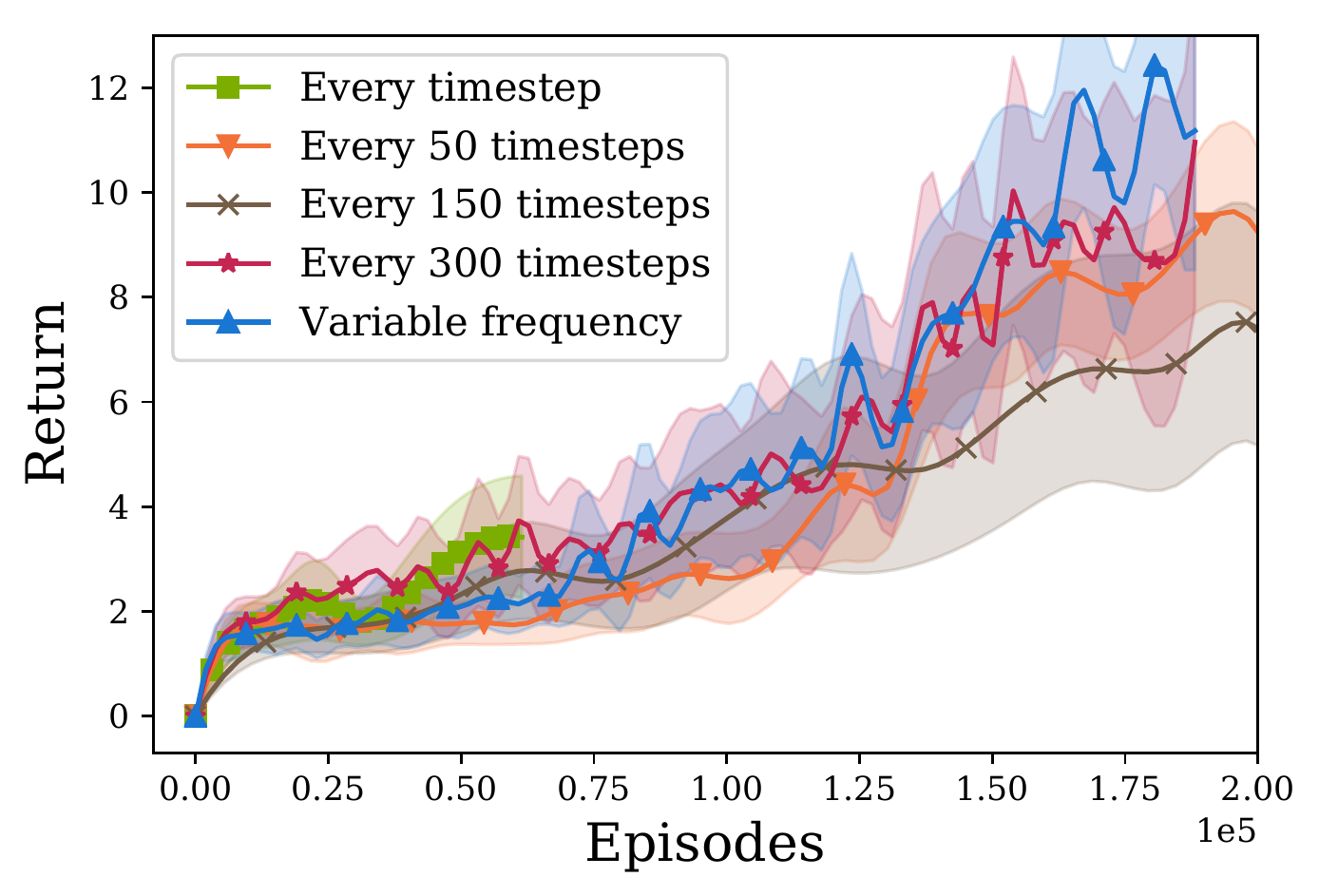}
%   \caption{\parbox[t]{.95\linewidth}{\centering Effect on performance.}}
%   \label{fig:freqComparison_reward}
\end{subfigure}\hfill%
\begin{subfigure}[t]{.6\columnwidth}
\hspace{-20pt}
%   \centering
        \begin{tabular}{@{}llll@{}}
        \toprule
        \textbf{\begin{tabular}[c]{@{}l@{}}High level \\ frequency\\ (every n timesteps)\end{tabular}} & \textbf{\begin{tabular}[c]{@{}l@{}}Inference \\ time (s)\end{tabular}} & \textbf{\begin{tabular}[c]{@{}l@{}}Effective\\ policy size\end{tabular}} & \textbf{\begin{tabular}[c]{@{}l@{}}Training\\ speed\\ (timesteps/s)\end{tabular}} \\ \midrule
        1                                                                                              & 0.03000                                                                & 3256                                                                     & 210.2                                                                             \\
        50                                                                                             & 0.00070                                                                & 359.1                                                                    & 1977.3                                                                            \\
        150                                                                                            & 0.00030                                                                & 319.7                                                                    & 1689.8                                                                            \\
        300                                                                                            & 0.00020                                                                & 309.8                                                                    & 1803.8                                                                            \\
        Variable freq.                                                                                 & 0.00055                                                                & 328.2                                                                    & 1774.2                                                                            \\ \bottomrule
        \end{tabular}
  \label{fig:freqComparison_time}
\end{subfigure}
\caption{Comparing hierarchical policies with HL running at different frequencies. 
% Variable frequency corresponds to the case where LL running duration is learned and set by the high-level. This is compared with HL running at different fixed frequencies. 
Left: Running HL at lower frequencies has minimal effect on performance. Right: Without temporal abstraction, the inference and training speed is low.}
\label{fig:freqComparison}
% \vspace{1cm}
\end{figure}

\textbf{Impact of High Level Frequency on Training and Inference.} To study the effect of temporal abstraction, we compare policies with different HL execution durations, trained on the goal finding task (see Fig.~\ref{fig:freqComparison}). % and Table~\ref{tab:freqEval}.
On the left, we show the learning curves for policies with the HL running once every $1$, $50$, $150$, $300$, and $d$ time-steps, where $d$ is the variable time interval output by the HL. We see that all variants are able to learn the task with minor differences in performance. However, notice that HL running at every time-step has made comparatively little progress. 
This difference in training speed is captured in Column $4$ of the table on the right.
It is clearly inefficient to run the HL every time given that we can achieve the optimal return much faster with the temporally abstracted policies. The exact inference times are recorded in Column $2$. Inference on temporally abstracted policies is $\approx 100$ times faster, which will also facilitate deployment on hardware. Column~$3$ calculates the effective size of the policies over time due to variation in HL frequency.

\section{Discussion and Conclusion}
We presented an HRL technique to solve visual navigation for a quadruped from pixels to leg motions. 
Our method outperformed non-hierarchical baselines on $3$ navigation tasks and achieved higher data efficiency and a lower wall-clock training time. 
However, the advantages of using our approach extend beyond performance improvements. 
First, by decoupling the high level from the low level, we were able to run them at different frequencies. Indeed, the high level learns when it has to process a new image. This has practical applications as processing vision input synchronously with the low level control loop is often impractical.
Secondly, we analyzed low level policies and demonstrated that they can be transferred between tasks. This is important as it is non-trivial to define 
skills - encoded by the latent command space - that are both robust and exploit the full range of robot capabilities. Transfer of low level policies makes it possible to use the learned skills as a continuous low-level action space for other learning algorithms, a research direction we intend to pursue in future work.

\textbf{Note on Hardware Evaluation.}
Due to COVID-19 restrictions, we were unable to include hardware results. However, we are planning to implement our hierarchical policies on a real Laikago robot once we regain access to our lab spaces. We have previously validated learned low-level policies similar to those found by our algorithm on multiple real robots, and we are therefore confident in the transfer of our HRL policies to hardware. We have also already tested our vision stack in combination with a predefined set of low-level skills on a legged robot. Finally, we found that similar latent commands were computed by the high level network when presented with real depth camera images and simulated ones in an environment with obstacles. More details can be found in Appendix~\ref{sec:robot_transfer}.

% \clearpage
% \input{sections/robot_transfer}

%===============================================================================

% The maximum paper length is 8 pages excluding references and acknowledgements, and 10 pages including references and acknowledgements
\clearpage
% The acknowledgments are automatically included only in the final version of the paper.
% \acknowledgments{If a paper is accepted, the final camera-ready version will (and probably should) include acknowledgments. All acknowledgments go at the end of the paper, including thanks to reviewers who gave useful comments, to colleagues who contributed to the ideas, and to funding agencies and corporate sponsors that provided financial support.}

%===============================================================================

\bibliography{references}  % .bib
\clearpage
\appendix
\section{Feasibility of Deploying Our Learned Policies on the Robot}
\label{sec:robot_transfer}
\begin{figure}[h]
\includegraphics[trim=0 80 0 0,clip,width=0.9\linewidth]{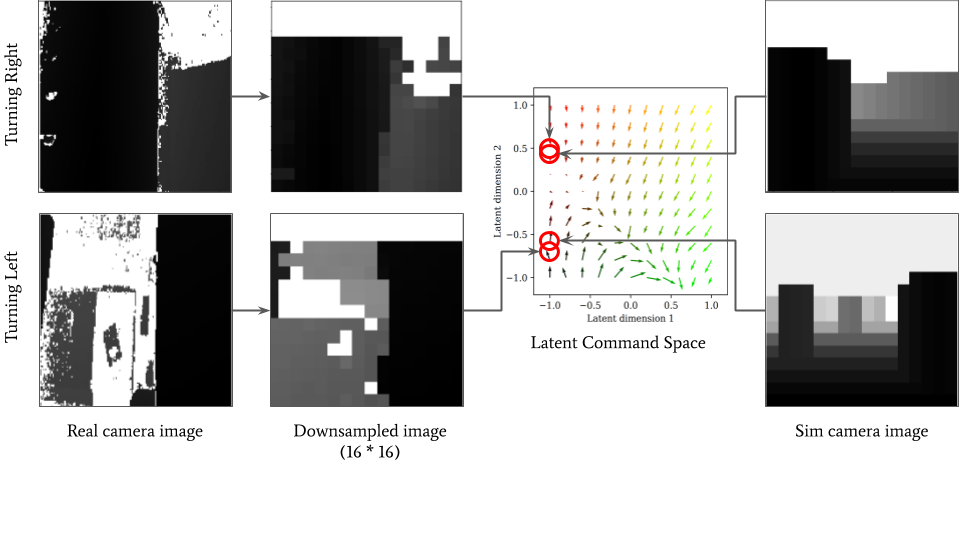}
\caption{Real depth camera images processed by high level.}
\label{fig:HLRobotTransfer}
\end{figure}

% In Fig.~\ref{fig:HLRobotTransfer}, we show that 
% The high level can process real depth camera images to issue latent commands (Fig.~\ref{fig:HLRobotTransfer}). 
We trained a hierarchical policy on goal finding task in simulation and
% HL input was generated from a simulated depth camera. 
% Sample HL from simulated camera are shown in Fig.~\ref{fig:HLRobotTransfer}. The corresponding latent commands output by HL are marked on the learned latent command space.
 evaluated the learned HL on images from a real depth camera (Intel RealSense L515).
We compared the downsampled ($16 \times 16$) real world camera images with similar looking simulated camera images from our experiments. 
% The resolution of real images was reduced to $16 \times 16$ to match that of simulated images. 
Both types of images result in similar latent commands, supporting compatibility of HL with real depth camera images (Fig.~\ref{fig:HLRobotTransfer}). 
% Top row in Fig.~\ref{fig:HLRobotTransfer} shows an example of selecting latent command corresponding to turning left when an obstacle is detected to the right in both simulated and real images. Bottom row shows a case for turning right.

\begin{figure}[h]
\vspace{-0.04in}
\captionsetup{singlelinecheck = false, format= hang, justification=raggedright}
\begin{overpic}[trim=30 0 60 150,clip,width=0.175\linewidth]{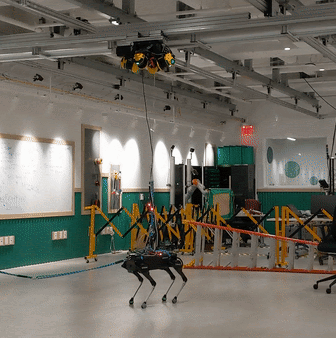}
 \put (10,10) {0.5s}
\end{overpic}%
\begin{overpic}[trim=30 0 60 150,clip,width=0.175\linewidth]{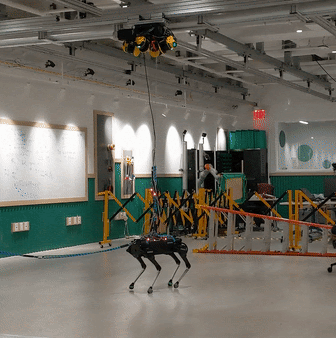}
 \put (10,10) {1.5s}
\end{overpic}%
\begin{overpic}[trim=30 0 60 150,clip,width=0.175\linewidth]{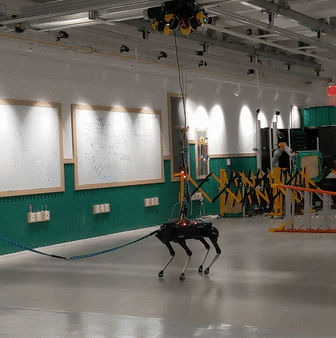}
 \put (10,10) {3.5s}
\end{overpic}%
\begin{overpic}[trim=30 0 60 150,clip,width=0.175\linewidth]{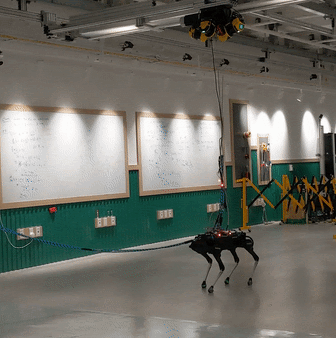}
 \put (10,10) {4.5s}
\end{overpic}
% \begin{overpic}[trim=30 0 60 150,clip,width=0.2\linewidth]{figures/demoday/frame_158_delay-0.04s.png}
%  \put (10,10) {6.5s}
% \end{overpic}
\caption{Forward walking policy deployed on hardware.}
\label{fig:demoDay}
\vspace{-0.1in}
\end{figure}
\begin{wrapfigure}[4]{R}{0.3\linewidth}
\vspace{-1.5in}
\centering
\includegraphics[width=.95\linewidth]{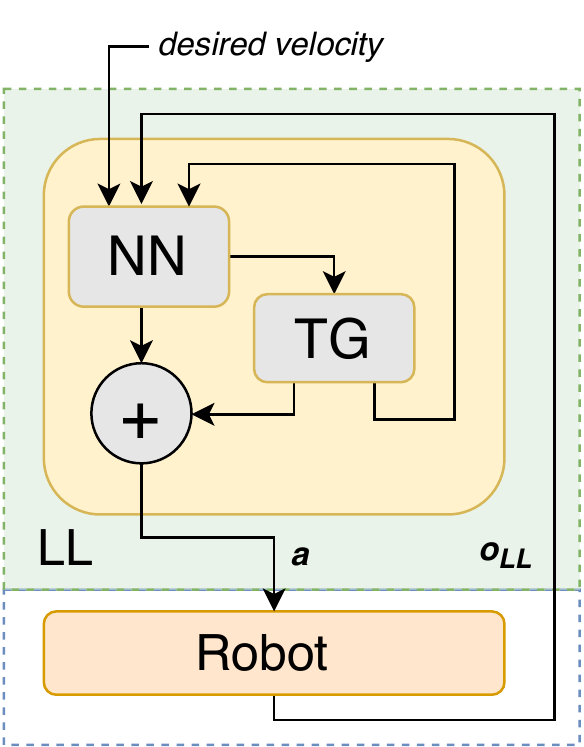}
% \caption{Training curves for hierarchical policies with $130K$ parameters.}
\label{fig:ll_schema}
\end{wrapfigure}

We have previously deployed a forward walking policy (trained in simulation) that tracks a desired velocity to the real Laikago robot (Fig.~\ref{fig:demoDay}). The hierarchical policies presented in our experiments are trained with a similar infrastructure. Hence, we expect that our policies will transfer to hardware. 

\begin{figure}[h]
\begin{overpic}[trim=150 75 150 20,clip,width=0.16\linewidth]{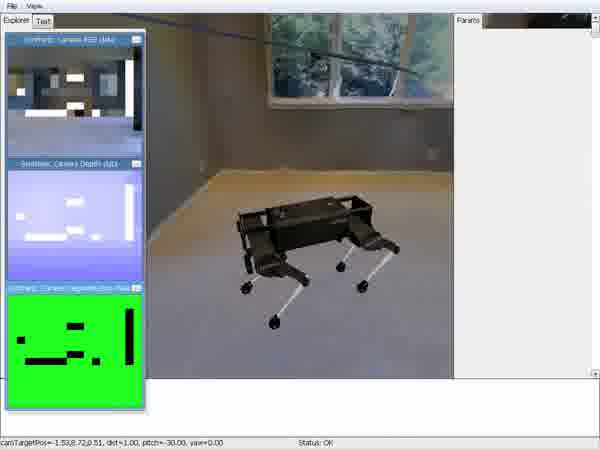}
 \put (10,10) {4s}
\end{overpic}%
\begin{overpic}[trim=150 75 150 20,clip,width=0.16\linewidth]{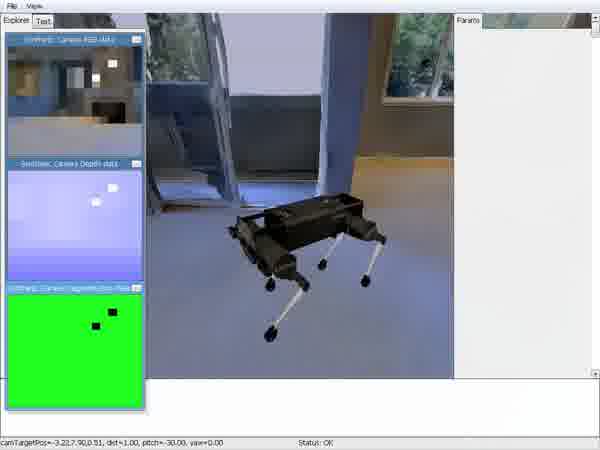}
 \put (10,10) {6s}
\end{overpic}%
\begin{overpic}[trim=150 75 150 20,clip,width=0.16\linewidth]{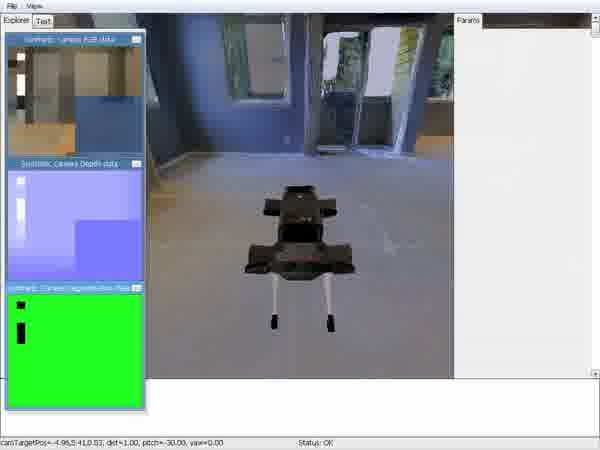}
 \put (10,10) {11s}
\end{overpic}%
\begin{overpic}[trim=150 75 150 20,clip,width=0.16\linewidth]{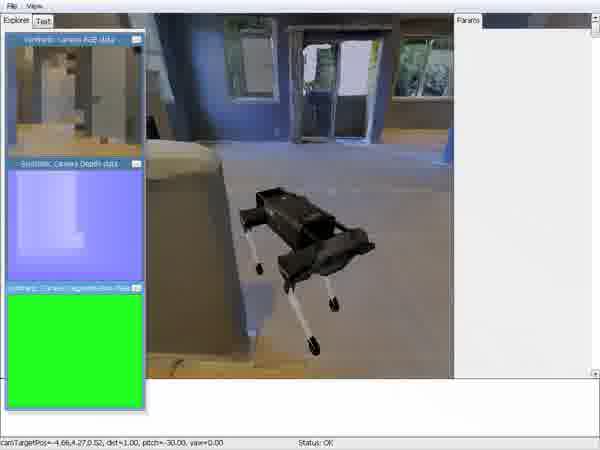}
 \put (10,10) {12s}
\end{overpic}%
\begin{overpic}[trim=150 75 150 20,clip,width=0.16\linewidth]{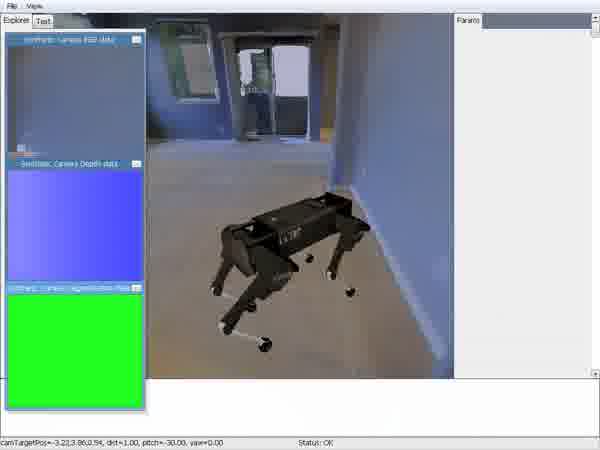}
 \put (2,10) {16s}
\end{overpic}%
\begin{overpic}[trim=150 75 150 20,clip,width=0.16\linewidth]{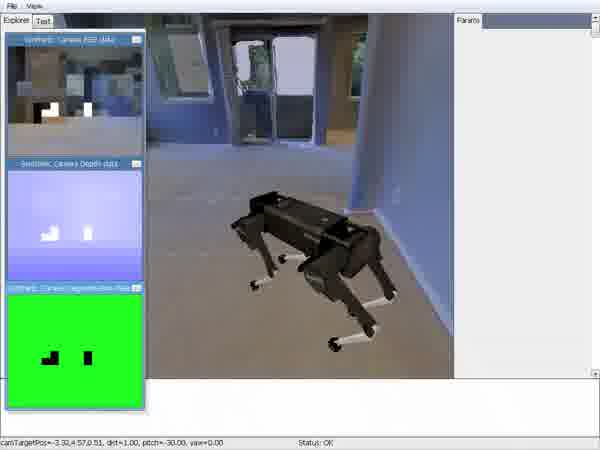}
 \put (10,10) {18s}
\end{overpic}\\
\begin{overpic}[trim=150 75 150 20,clip,width=0.16\linewidth]{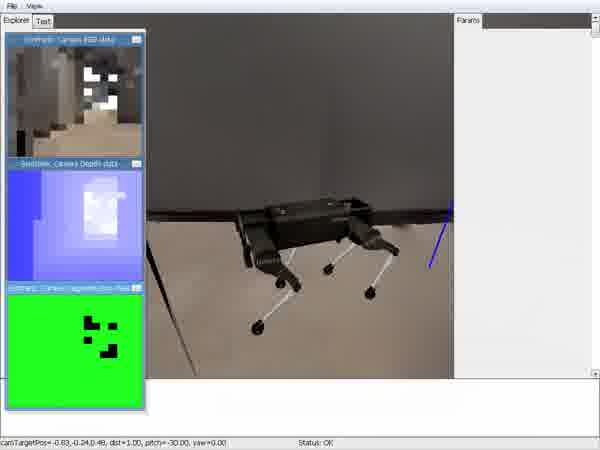}
 \put (10,10) {4s}
\end{overpic}%
\begin{overpic}[trim=150 75 150 20,clip,width=0.16\linewidth]{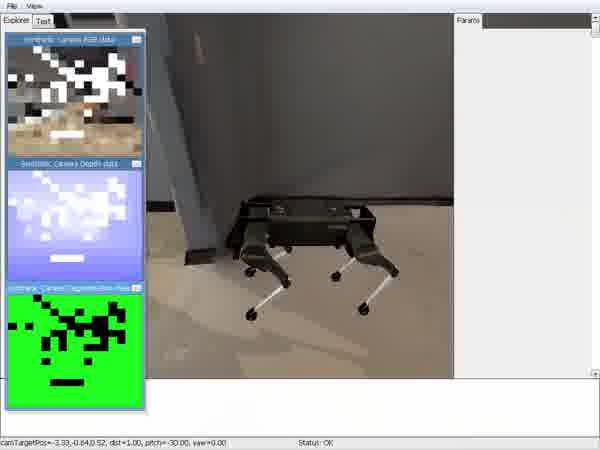}
 \put (10,10) {8s}
\end{overpic}%
\begin{overpic}[trim=150 75 150 20,clip,width=0.16\linewidth]{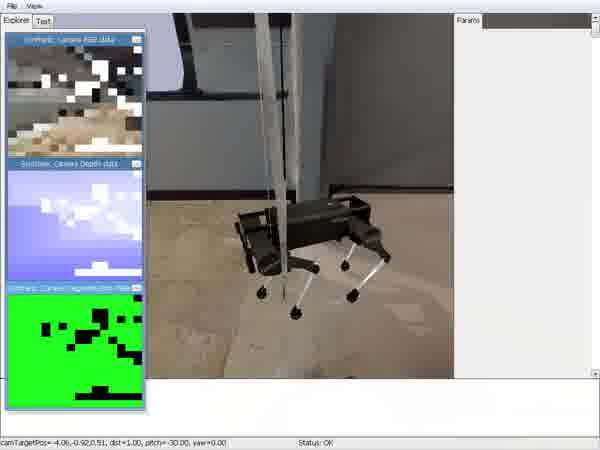}
 \put (10,10) {9s}
\end{overpic}%
\begin{overpic}[trim=150 75 150 20,clip,width=0.16\linewidth]{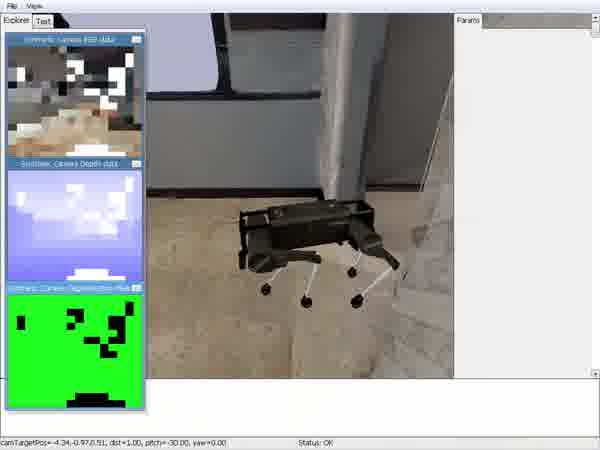}
 \put (10,10) {10s}
\end{overpic}%
\begin{overpic}[trim=150 75 150 20,clip,width=0.16\linewidth]{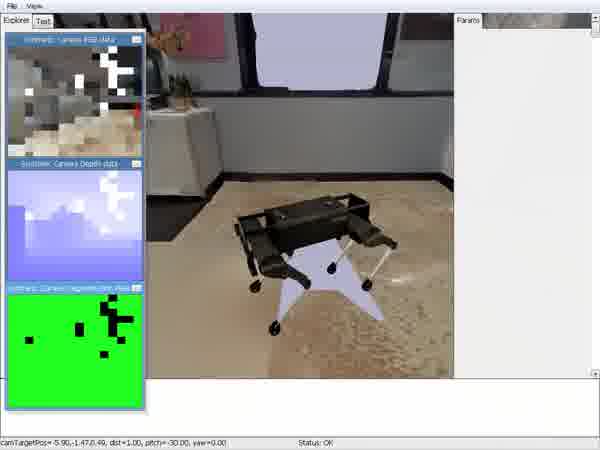}
 \put (10,10) {13s}
\end{overpic}%
\begin{overpic}[trim=150 75 150 20,clip,width=0.16\linewidth]{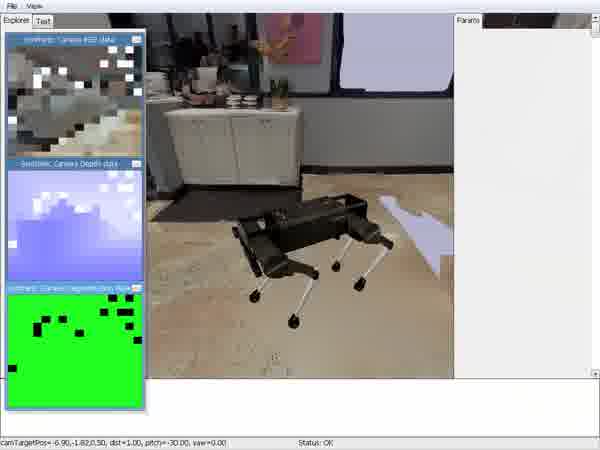}
 \put (10,10) {15s}
\end{overpic}
\caption{Navigating Gibson environments with hierarchical policies.}
\label{fig:gibson}
\end{figure}

Finally, we trained our policies in simulated 3D spaces with realistic visuals from the Gibson dataset~\cite{xiazamirhe2018gibsonenv}. After training, our policies were able to transfer to a new space (Fig.~\ref{fig:gibson}).

% Recordings of our policies are presented in the supplementary video.

\end{document}